\documentclass[letterpaper,journal]{IEEEtran}

\usepackage{amsmath,amssymb,amsfonts}
\usepackage{algorithmic}
\usepackage{graphicx}
\usepackage{textcomp}
\usepackage{xcolor}
\usepackage{booktabs}
\usepackage{multirow}
\usepackage{cite}
\usepackage{hyperref}
\usepackage[caption=false]{subfig}
\usepackage{subcaption}
\usepackage{algorithm}
\usepackage{tikz}
\usepackage{pgfplots}
\pgfplotsset{compat=1.17}
\usepackage{enumitem}
\usepackage{pifont}
\usepackage{multirow}
\usepackage{makecell}
\usepgfplotslibrary{groupplots}
\usetikzlibrary{arrows.meta, positioning, shapes, fit}
\newcommand{\std}[1]{\,{\scriptsize $\pm$#1}}
  
\usetikzlibrary{patterns} 
\definecolor{naiveA}{RGB}{ 44,111,172} 
\definecolor{naiveB}{RGB}{ 76,133,186} 
\definecolor{naiveC}{RGB}{108,155,200} 
\definecolor{naiveD}{RGB}{140,178,215} 
\definecolor{naiveE}{RGB}{172,200,229} 
\definecolor{lutA}{RGB}{185, 64, 64} 
\definecolor{lutB}{RGB}{199, 96, 96} 
\definecolor{lutC}{RGB}{213,128,128} 
\definecolor{lutD}{RGB}{226,160,160} 
\definecolor{lutE}{RGB}{240,192,192}

\newcommand{\review}[1]{{\color{black}#1}}

\usepackage{xspace}
\newcommand{\sysname}{\textbf{Bern2Edge}\xspace}

\usepackage{titlesec}
\titlespacing*{\section}{0pt}{6pt}{4pt}
\titlespacing*{\subsection}{0pt}{4pt}{2pt}

\begin{document}

\title{\sysname: A Neurosymbolic Compiler for Edge Deployment via Bernstein Polynomial Networks
\thanks{\review{Code repo can be found at: \url{https://github.com/PervasiveAutonomyLab/Bern2Edge}. Archived at \url{https://zenodo.org/records/21726441}.}}}

\author{Malak~Gamal~El-Din,~Yifan~Zhang,~Yasser~Shoukry,~Sitao~Huang,~Salma~Elmalaki
\thanks{All authors are with the Department of Electrical Engineering and Computer Science, University of California, Irvine, CA 92697 USA (e-mail: mgamalel@uci.edu; yifanz58@uci.edu; yshoukry@uci.edu; sitaoh@uci.edu; selmalak@uci.edu).}}

% \thanks{\scriptsize
% \copyright~2026 IEEE. Personal use of this material is permitted.
% Permission from IEEE must be obtained for all other uses, in any current
% or future media, including reprinting/republishing this material for
% advertising or promotional purposes, creating new collective works, for
% resale or redistribution to servers or lists, or reuse of any copyrighted
% component of this work in other works.}}

\IEEEpubid{%
\parbox{\textwidth}{%
\vspace{10pt}
\centering\scriptsize
\copyright~2026 IEEE. Personal use of this material is permitted.
Permission from IEEE must be obtained for all other uses, in any current
or future media, including reprinting/republishing this material for
advertising or promotional purposes, creating new collective works, for
resale or redistribution to servers or lists, or reuse of any copyrighted
component of this work in other works.
}}

\maketitle

% \vspace{-80pt}

\begin{abstract}
Deploying high-accuracy neural networks on resource-constrained edge
devices remains challenging, as existing approaches treat training,
compression, and hardware synthesis as separate stages, leaving a gap
between software-trained models and efficient end-to-end deployment with
limited support for interpretability. We propose \sysname, an end-to-end
framework that uses knowledge distillation to convert a pretrained teacher
feed-forward network into hardware-efficient representations via Bernstein polynomial
activations. This representation enables two deployment paths: (i) a high-fidelity
LUT-based realization that preserves model fidelity under compression, and
(ii) a symbolic rule-based representation derived from Bernstein
activation geometry, enabling interpretable inference with explicit input-space constraints. The resulting BNNs achieve
up to $2.12$ percentage-point (pp) accuracy improvement over ReLU under
identical compression constraints. At the system level, \sysname achieves
up to $99.8$\% latency reduction and $95.2$\% BRAM reduction relative to a
W8A8 quantized teacher on an AMD Xilinx KV260 FPGA, while maintaining
accuracy within $0.5$\,pp, and further deploys on a low-power Spartan-7
XC7S15 FPGA. The rule-based path reduces DSP usage by up to $89.0$\% at a
cost of $1.5$\,pp in total accuracy.
\end{abstract}

\begin{IEEEkeywords}
Bernstein polynomials, activation functions, knowledge distillation,
lookup table, FPGA inference, symbolic rule extraction, edge ML.
\end{IEEEkeywords}

\section{Introduction}
The rapid growth of deep neural networks (DNNs) has driven significant gains in predictive performance, but has made deployment on resource-constrained edge devices increasingly difficult. Networks are typically trained in software without accounting for hardware constraints, yielding models that are hard to map to strict latency and resource budgets, while limited interpretability further restricts their use in edge settings where transparency and reliability are critical. Bridging high-level model design and efficient hardware realization remains a key challenge, as existing approaches treat model design and hardware optimization separately, limiting their ability to satisfy both accuracy and system-level constraints~\cite{jiang2020hardware}.

Prior work addresses parts of this problem through model compression, quantization, and hardware compilation frameworks such as hls4ml~\cite{duarte2018hls4ml}, FINN~\cite{blott2018finn}, and CGRA4ML~\cite{cgra4ml2026}, which translate trained networks into synthesizable hardware, often applying quantization and pruning during deployment. However, these operate on fixed neural representations, mapping or optimizing pre-trained models for hardware rather than jointly designing representations inherently aligned with hardware efficiency.

We introduce \sysname, an end-to-end neurosymbolic compiler for edge deployment based on Bernstein polynomial activations~\cite{khedr2024deepbern}, targeting feed-forward networks (FFNs), including multilayer perceptrons (MLPs) and the FFN sublayers of transformers. The structured, bounded representation of Bernstein polynomials suits both hardware realization and symbolic reasoning~\cite{farouki2012bernstein}. Our framework transforms a trained DNN teacher into hardware-efficient representations via knowledge distillation (KD)~\cite{hinton2015distilling}, training a compact student with Bernstein activations, which we refer to as a Bernstein Neural Network (BNN). This supports two deployment paths: a lookup table (LUT)-based realization, where a per-neuron LUT represents the Bernstein activation for compact, efficient synthesis, and an optional symbolic rule-based representation derived from Bernstein activation geometry, enabling further compression and interpretable inference. To our knowledge, this is the first work to jointly exploit the structural properties of Bernstein polynomial activations for both efficient hardware synthesis and symbolic rule extraction.
\IEEEpubidadjcol
\textbf{The contributions of this work are as follows:}
\begin{itemize}[topsep=0pt]
\item \textbf{End-to-end teacher-to-hardware deployment framework.}
We propose \sysname, a unified pipeline compressing a trained teacher DNN into a hardware-efficient BNN via KD, supporting two deployment paths: a LUT-based realization for efficient hardware execution, and a symbolic rule-based representation for additional compression and interpretable inference.
\item \textbf{Bernstein activations for improved compression under knowledge distillation.}
We introduce Bernstein polynomial activations as a structured student representation that improves KD under strong compression: BNNs recover the accuracy of large DNN teachers more effectively than standard activations such as ReLU under compressed regimes, while remaining well-suited for hardware realization and interpretability.
\item \textbf{LUT-based hardware realization.}
Bernstein polynomial activations admit a direct LUT-based implementation leveraging their normalized input domain and fixed functional structure, enabling efficient per-neuron LUT realization and a compact representation for edge inference.
\item \textbf{Symbolic rule extraction and synthesis.}
We develop a BNN-specific rule extraction method deriving compact symbolic rules from the geometric structure of learned Bernstein activations, forming a structured, hardware-efficient representation that supports quantization to low bitwidth with negligible accuracy loss.
\item \textbf{Evaluation in compression and deployment regimes.}
We evaluate \sysname on multiple tabular datasets and a transformer FFN setting, demonstrating: (i) improved performance of BNNs under strong compression, (ii) efficient LUT-based hardware synthesis at cost comparable to or less than standard activations, (iii) compact and effective symbolic rule representations, (iv) end-to-end deployment on AMD Xilinx KV260 and low-power Spartan-7 XC7S15 FPGAs, and (v) robustness of the rule network under input noise and distribution shift.
\end{itemize}

\section{Background and Related Work}
\label{sec:related}

\subsection{Edge Deployment Frameworks and Model Compression}
A large body of work targets neural network deployment on resource-constrained hardware through compiler frameworks and model compression. hls4ml~\cite{duarte2018hls4ml} translates trained networks into FPGA implementations via high-level synthesis, mapping dense and convolutional operators to streaming architectures with fixed-point quantization. FINN~\cite{blott2018finn} extends this to quantized networks, generating dataflow accelerators for low-bitwidth, high-throughput FPGA inference. CGRA4ML~\cite{cgra4ml2026} extends these approaches to coarse-grained reconfigurable architectures (CGRAs). These systems primarily map and optimize pre-trained models for hardware through post-training transformations (quantization, pruning, operator fusion) rather than co-designing the model representation itself.

Our work instead targets \emph{training-time} representation: we train Bernstein polynomial activations whose learned forms are inherently deployable, rather than adapting arbitrary activations for hardware after the fact, enabling direct symbolic extraction and a unified pipeline from training to hardware implementation.

Since many high-performing models are pretrained with ReLU-based architectures, we employ knowledge distillation (KD) to transfer large pretrained teachers into compact BNNs. KD is a standard compression technique matching soft outputs or intermediate representations between a teacher and student~\cite{hinton2015distilling}. In \sysname, KD trains student BNNs that retain the accuracy of much larger models while maintaining a structured, hardware-friendly representation.

\subsection{Hardware Realization of Nonlinear Activations}
Efficient implementation of nonlinear activations remains a key challenge in hardware deployment. Piecewise-linear functions such as ReLU map efficiently to comparator-based logic, but expressive smooth activations such as GeLU~\cite{hendrycks2016gelu} and Swish~\cite{ramachandran2017swish} require more complex arithmetic and are typically approximated using lookup tables or piecewise-polynomial methods~\cite{choi2024swish_hw}, introducing a train--deploy gap between the trained and deployed function.

LUT-centric designs address this by mapping neuron computations directly to lookup tables. LUTNet~\cite{wang2019lutnet} uses native FPGA LUTs as inference operators to reduce arithmetic cost, while LogicNets~\cite{umuroglu2020logicnets} co-designs sparse, quantized neurons extractable as LUT truth tables. Polynomial-based methods such as PolyLUT~\cite{andronic2023polylut} and PolyLUT-Add~\cite{lou2024polylutadd} extend this to structured polynomial classes, but both require hard fan-in limits of $\leq 7$ inputs per neuron to bound LUT size: PolyLUT enforces sparse connectivity, while PolyLUT-Add sums low-fan-in sub-neurons. This is incompatible with the dense tabular MLPs we target ($14$--$54$ input features), requiring additional sparsification or redesign to apply. \sysname instead retains dense MLP layers and makes the learned activation itself hardware-realizable.

KANEL\'E~\cite{kanele2025} explores activation-centric LUT mappings via Kolmogorov--Arnold Networks (KANs), replacing MLP computations with learnable one-dimensional edge functions discretized into per-neuron LUTs, departing from standard MLP structure. In contrast, \sysname retains standard FFN architectures and targets the activation directly: Bernstein activation coefficients define the deployed function, enabling exact realization via small per-neuron LUTs without a train--deploy gap.

\subsection{Bernstein Polynomial Activations}

\noindent \textbf{$\bullet$ Bernstein Polynomials.}
A Bernstein polynomial $\sigma(z)$ of degree $n$, defined on the interval $[l, u]$, is written as~\cite{farouki2012bernstein}:
\begin{equation}
\label{algo:BernAct}
  \sigma(z;\, l, u, \mathbf{c}) = \sum_{k=0}^{n} c_k \, b_{n,k}(t),
  \quad t = \frac{z - l}{u - l} \in [0,1], \quad z \in [l, u],
\end{equation}
where $z$ is the input to the polynomial function, $t$ is the normalized input, and $\mathbf{c} = \{c_0, \ldots, c_n\}$ are the polynomial coefficients controlling the activation shape. The Bernstein basis functions are defined as:
\begin{equation}
  b_{n,k}(t) = \binom{n}{k} t^k (1 - t)^{n-k}, \quad k = 0, \ldots, n,
\end{equation}
satisfying the partition of unity property:
\begin{equation}\label{eq:unity}
  \sum_{k=0}^{n} b_{n,k}(t) = 1, \quad \forall\, t \in [0,1].
\end{equation}
This defines the polynomial as a convex combination of its coefficients over a bounded domain, with the coefficients directly controlling the activation shape.

\noindent \textbf{$\bullet$ Bernstein Neural Networks (BNNs).}
Bernstein polynomials have been explored in neural networks as a nonlinear activation due to their favorable analytic and approximation properties. Khedr et al.~\cite{khedr2024deepbern} introduce Deep Bernstein Networks, feed-forward networks in which standard activations are replaced by learnable Bernstein polynomials. We adopt this formulation throughout \sysname and refer to it as a Bernstein Neural Network (BNN). For a network of depth $L$, the input is denoted $\mathbf{x}^{(0)}$ and the output of the $l$-th layer $\mathbf{x}^{(l)}$, with propagation rule:
\begin{equation}\label{eq:propagation}
\mathbf{x}^{(l)} = \sigma\left( \mathbf{z}^{(l)}; \mathbf{c}^{(l)} \right), \quad z_i^{(l)} = \left(\mathbf{w}_i^{(l)}\right)^{\!\top} \mathbf{x}^{(l-1)} + \beta_i^{(l)}
\end{equation}
where $\mathbf{w}_i^{(l)}$ and $\beta_i^{(l)}$ are the learnable weights and biases. The activation function $\sigma$ operates element-wise, parametrized by learnable Bernstein coefficients $\mathbf{c}^{(l)} = \{c_{k}^{(l)}\}_{k=0}^{n}$, where $n$ is a hyperparameter for the polynomial degree, allowing the network to learn the shape of its nonlinearities alongside its weights. For simplicity of notation, we drop the superscript $(l)$ in the remainder of the paper.

\noindent \textbf{$\bullet$ Theoretical Properties of BNNs.}
(1) Stable training: unlike other polynomial activations, which suffer exploding-gradient instability that worsens with degree~\cite{goyal2020improved}, Bernstein polynomials remain stable~\cite{khedr2024deepbern, albool2026deepbern}.
(2) Exponential approximation rates and parameter efficiency: Bernstein-based architectures achieve approximation error bounds superior to standard feed-forward ReLU networks~\cite{albool2026deepbern}, approximating smooth functions at much lower complexity than piecewise-linear ReLU networks. These properties motivate the use of BNNs to compress larger networks.

In \sysname, we exploit the structure of the Bernstein polynomial activation for both hardware synthesis and symbolic rule extraction.

\subsection{Symbolic Rule Extraction}
Prior work on symbolic rule extraction from neural networks falls into pedagogical and decompositional methods. Pedagogical approaches such as TREPAN~\cite{craven1995trepan} treat the trained network as a black box and induce a decision tree by querying its input--output behavior. Decompositional approaches instead exploit internal network structure: DeepRED~\cite{zilke2016deepred} extends earlier rule extraction methods to deep networks, extracting intermediate rules layer by layer and using decision trees to describe hidden-layer behavior, while ECLAIRE~\cite{zarlenga2021eclaire} improves the scalability of decompositional rule extraction while maintaining high-quality, interpretable rule sets.  NeuSym-HLS~\cite{pan2025neusym} applies hardware-aware symbolic regression to replace internal neural layers with compact analytic expressions.

Earlier activation-aware techniques such as validity interval analysis~\cite{thrun1994rules} propagate activation intervals through the network to derive symbolic descriptions, but treat activations generically without exploiting a learned activation family's analytic structure.

\sysname differs from these approaches by deriving rule candidates directly from the geometric structure of the learned Bernstein activations, instead of inducing decision trees or propagating interval bounds. This yields semantically aligned partitions tied to the network's learned nonlinear representation and supports a compact symbolic rule set suitable for efficient deployment.

\section{\sysname Framework Overview}
\label{sec:framework}

\begin{figure*}[t]
    \centering
    \includegraphics[
        width=2\columnwidth, % left bottom right top
            trim={6cm 0 12cm 0}, % left bottom right top
    clip
    ]{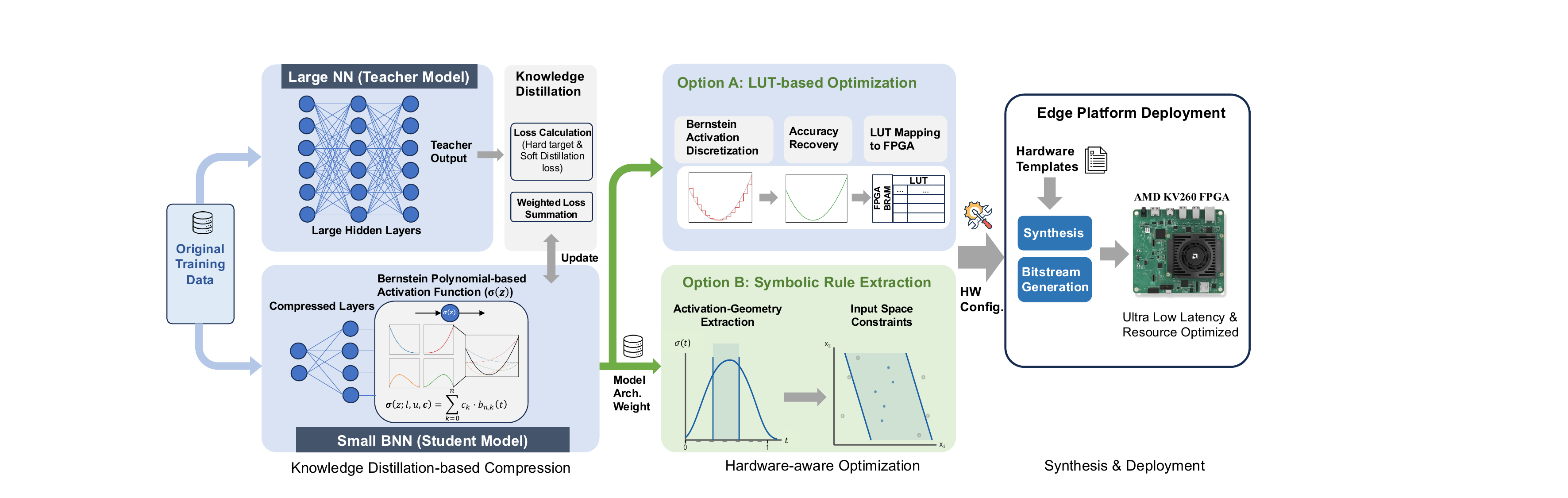}
    \caption{Overview of \sysname: A high-accuracy teacher model is distilled into a compressed BNN student via KD. The resulting representation is synthesized and deployed via either exact LUT-based realization or symbolic rule extraction.}
    \label{fig:framework_pipeline}
\end{figure*}

We propose \sysname, an end-to-end pipeline that transforms a trained teacher model into hardware-efficient and interpretable representations. The framework consists of four modules: (1) BNN training via KD, followed by two alternative deployment paths: (2) LUT-based hardware realization, or (3) symbolic rule extraction, both followed by (4) hardware synthesis and deployment. Fig.~\ref{fig:framework_pipeline} shows the \sysname pipeline. The following sections describe each module and the Bernstein polynomial properties that enable these capabilities.

% ============================================================
\subsection{BNN Training via Knowledge Distillation}
\label{sec:training}
% ============================================================
\noindent\textbf{Knowledge Distillation Setup.}
The first step in the \sysname pipeline is training compressed BNNs as student models using KD~\cite{hinton2015distilling}, reducing model size while preserving accuracy.

Bernstein-based networks exhibit favorable approximation properties under a constrained parameter budget, enabling accurate function representation with fewer parameters than standard activations~\cite{albool2026deepbern}, motivating our use of Bernstein activations in the student network: a Bernstein-based student can recover more of the teacher's accuracy than a ReLU-based student of equal size, an effect confirmed across all evaluated architectures in Section~\ref{sec:results_kd}.

\begin{figure}[t]
  \centering
    \includegraphics[width=1\columnwidth]
    {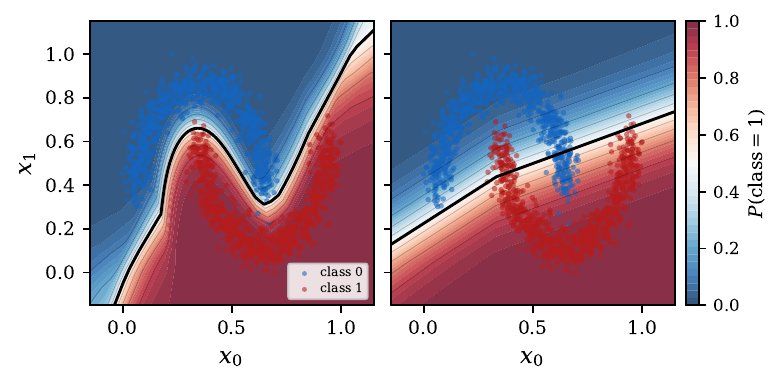}
  \caption{Decision boundaries learned by a BNN (left) and a ReLU model (right) of equal capacity on the synthetic Two Moons dataset~\cite{pedregosa2011scikit}. The BNN captures the curved structure of the data, while the ReLU model produces a simpler, predominantly linear boundary.}
  \label{fig:decision_boundary}
\end{figure}

This is further reflected in the BNN's learned decision geometry: Bernstein-based models capture smoother, more curved decision boundaries, while standard activations such as ReLU tend to produce piecewise-linear boundaries, as illustrated in Fig.~\ref{fig:decision_boundary}.

The student objective follows the standard distillation loss~\cite{hinton2015distilling}:
\begin{equation}
\mathcal{L}_s=(1-\alpha)\mathcal{L}_{\mathrm{CE}}
+\alpha T^2\,\mathrm{KL}\!\left(
S(\mathbf{y}^*/T)\,\middle\|\,S(\hat{\mathbf{y}}/T)
\right),
\label{eq:kd}
\end{equation}
where $\mathcal{L}_{\mathrm{CE}}$ is the cross-entropy loss between student predictions and ground-truth labels, $S(\cdot)$ is the softmax, $\mathrm{KL}(\cdot\|\cdot)$ is the Kullback--Leibler divergence between softened teacher and student distributions, and $\hat{\mathbf{y}}$ and $\mathbf{y}^*$ are the student and teacher logits, with $\hat{\mathbf{y}} = \mathbf{x}^{(L)}$ the student's final layer output. $T$ is the temperature controlling output-distribution softness, and $\alpha$ balances supervision between hard labels and teacher guidance.

Teacher models are selected to provide strong supervision signals consistent with prior tabular benchmarks, following the MLP baselines of~\cite{gorishniy2021}. Student architectures are designed to satisfy strict compression and latency constraints. While Bernstein activations introduce additional learnable parameters as polynomial coefficients, these do not translate to increased hardware cost under our LUT-based activation implementation, as described next (Section~\ref{sec:hardware}).

\noindent\textbf{BNN Fixed-Bound Training.}
Bernstein activations operate over a fixed input domain, requiring inputs normalized to $[0,1]$ as in Eq.~\ref{algo:BernAct}, a property later leveraged for efficient hardware realization. After a short warmup phase, per-neuron pre-activation bounds $[l_i, u_i]$ are calibrated from empirical activation statistics and fixed for the remainder of training, sequentially, one layer at a time, once preceding layers have stabilized.

Fixing these bounds serves two purposes: stabilizing training by preventing continual shifts in activation normalization, and enforcing the fixed-domain consistency required for exact LUT-based deployment (Section~\ref{sec:hardware}).

To discourage out-of-bounds inputs, we add an out-of-bounds penalty to the BNN loss function:
\begin{equation}\label{eq:studentpenalty}
 \mathcal{L}_{\mathrm{BNN}} = \mathcal{L}_s + \lambda_{\mathrm{ob}} \cdot \mathbb{E}\left[ 
  \max(0, -t) + \max(0, t - 1)
  \right],
\end{equation}
where $t$ denotes the normalized pre-activation, as in Eq.~\ref{algo:BernAct}. During forward propagation, activations are clamped after normalization to preserve the Bernstein basis structure, ensuring non-negativity and partition-of-unity as in Eq.~\ref{eq:unity}. Section~\ref{sec:results_kd} compares compressed student BNNs via KD against similarly compressed ReLU networks under identical hardware constraints.

\subsection{LUT-Based Hardware Realization}
\label{sec:hardware}
The next stage of the \sysname framework is the hardware synthesis of the trained BNN on FPGA. As shown in Fig.~\ref{fig:framework_pipeline}, one deployment path leverages a LUT-based realization of Bernstein activations.

Direct evaluation of the Bernstein activation in Eq.~\ref{algo:BernAct} is computationally expensive. For a polynomial of degree $n$, it requires $(n+1)$ power operations, $(n+1)$ multiplications with binomial coefficients, and an $(n+1)$-term dot product with learned coefficients, resulting in approximately $3(n+1)$ multiply-accumulate (MAC) operations per neuron. However, since Bernstein activations operate over a fixed normalized domain $[0,1]$, the activation function can be precomputed. Each neuron's activation is evaluated offline over a uniform grid of $E$ points in $[0,1]$ and stored in on-chip memory (BRAM). At runtime, activation evaluation reduces to a single memory access, eliminating all arithmetic operations.

BRAM storage scales linearly with the number of neurons $h$ in a layer and grid resolution $E$. While storage grows with $h$, the complete removal of activation-phase computation yields substantial savings in DSP usage and combinational logic. This trade-off motivates training of compact BNN students using KD (Section~\ref{sec:training}), keeping $h$ small enough for efficient on-chip storage.

A naive implementation uses nearest-neighbor lookup via floor indexing, which quantizes inputs into $E$ discrete levels and introduces an approximation error. To mitigate this, we employ low-cost linear interpolation between adjacent LUT entries:
\begin{equation}
f(t) = \mathcal{T}[i_{\mathrm{lo}}] + \mu \cdot \left(\mathcal{T}[i_{\mathrm{hi}}] - \mathcal{T}[i_{\mathrm{lo}}]\right),
\end{equation}
where $t$ is the normalized pre-activation as defined in Eq.~\ref{algo:BernAct}, $\mathcal{T}[\cdot]$ is the $E$-entry LUT, $i_{\mathrm{lo}}, i_{\mathrm{hi}}$ are the indices of the two adjacent grid points bracketing $t$, and $\mu \in [0,1]$ is the fractional position between them. This interpolation significantly improves accuracy while introducing minimal overhead: one multiply, one subtraction, and one addition per neuron, all implemented in combinational logic with no additional DSP or BRAM usage.

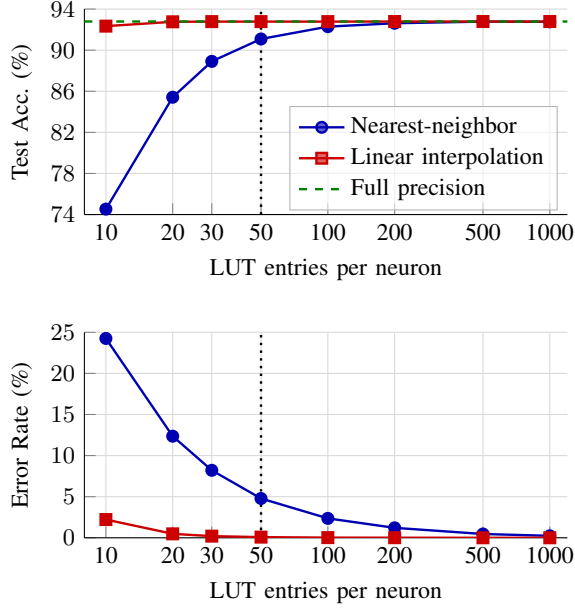
\begin{figure}[t]
  \centering
  \subfloat{\begin{tikzpicture}
\begin{axis}[
    width=0.9\columnwidth,
    height=4.3cm,
    xlabel={LUT entries per neuron},
    ylabel={Test Acc. (\%)},
    axis y line*=left,
    axis x line*=bottom,
    xmode=log,
    log basis x=10,
    xmin=8, xmax=1200,
    ymin=74, ymax=94,
    ytick={74,78,82,86,90,94},
    xtick={10,20,30,50,100,200,500,1000},
    xticklabels={10,20,30,50,100,200,500,1000},
    grid=both,
    grid style={line width=0.4pt, draw=gray!30},
    tick label style={font=\small},
    label style={font=\small},
    legend style={
        at={(0.98,0.02)},
        anchor=south east,
        font=\small,
        draw=gray!60,
        fill=white,
        fill opacity=0.6,
        text opacity=1,
    },
    legend cell align={left},
]

%% Nearest-neighbor accuracy
\addplot[
    color=blue!70!black,
    solid,
    mark=*,
    mark size=2pt,
    line width=0.9pt,
] coordinates {
    (10, 74.53)
    (20, 85.42)
    (30, 88.91)
    (50, 91.09)
    (100, 92.29)
    (200, 92.62)
    (500, 92.78)
    (1000, 92.76)
};
\addlegendentry{Nearest-neighbor}

%% Linear interpolation accuracy
\addplot[
    color=red!80!black,
    solid,
    mark=square*,
    mark size=2pt,
    line width=0.9pt,
] coordinates {
    (10, 92.34)
    (20, 92.76)
    (30, 92.78)
    (50, 92.78)
    (100, 92.78)
    (200, 92.79)
    (500, 92.79)
    (1000, 92.79)
};
\addlegendentry{Linear interpolation}

%% Full-precision baseline
\addplot[
    color=green!50!black,
    dashed,
    line width=0.9pt,
] coordinates {(8,92.79) (1200,92.79)};
\addlegendentry{Full precision}

%% Selected operating point: 50 entries
\addplot[
    color=black,
    dotted,
    line width=0.9pt,
    forget plot,
] coordinates {(50,74) (50,94)};

\end{axis}
\end{tikzpicture}}
  \hspace{0.01\columnwidth}
  \subfloat{\begin{tikzpicture}
\begin{axis}[
    width=0.9\columnwidth,
    height=4.3cm,
    xlabel={LUT entries per neuron},
    ylabel={Error Rate (\%)},
    axis y line*=left,
    axis x line*=bottom,
    xmode=log,
    log basis x=10,
    xmin=8, xmax=1200,
    ymin=0, ymax=25,
    ytick={0,5,10,15,20,25},
    xtick={10,20,30,50,100,200,500,1000},
    xticklabels={10,20,30,50,100,200,500,1000},
    grid=both,
    grid style={line width=0.4pt, draw=gray!30},
    tick label style={font=\small},
    label style={font=\small},
    legend style={
        at={(0.98,0.98)},
        anchor=north east,
        font=\small,
        draw=gray!60,
        fill=white,
        fill opacity=0.92,
        text opacity=1,
    },
    legend cell align={left},
]

%% Nearest-neighbor error
\addplot[
    color=blue!70!black,
    solid,
    mark=*,
    mark size=2pt,
    line width=0.9pt,
] coordinates {
    (10, 24.25)
    (20, 12.36)
    (30, 8.21)
    (50, 4.78)
    (100, 2.36)
    (200, 1.21)
    (500, 0.46)
    (1000, 0.24)
};

%% Linear interpolation error
\addplot[
    color=red!80!black,
    solid,
    mark=square*,
    mark size=2pt,
    line width=0.9pt,
] coordinates {
    (10, 2.22)
    (20, 0.48)
    (30, 0.20)
    (50, 0.08)
    (100, 0.02)
    (200, 0.01)
    (500, 0.00)
    (1000, 0.00)
};

%% Selected operating point: 50 entries
\addplot[
    color=black,
    dotted,
    line width=0.9pt,
    forget plot,
] coordinates {(50,0) (50,25)};

\end{axis}
\end{tikzpicture}}
  \caption{%
  Per-neuron LUT behavior vs.\ table size on Covertype: test accuracy (top) and error rate (bottom) vs.\ LUT entries per neuron. Error\% is the fraction of samples where LUT output differs from full-precision inference.
  }
  \label{fig:lut_resources}
\end{figure}

Empirically, for a BNN with hidden layer size $h=256$ on Covertype~\cite{blackard1999comparative}, linear interpolation at 50 entries (100\,KB BRAM) achieves an error rate of $0.08\%$, roughly $60\times$ lower than nearest-neighbor's $4.78\%$ at the identical table size and storage cost. We select 50 entries as our operating point for the remainder of this work. Nearest-neighbor error falls off slowly with table size and does not reach linear interpolation's 50-entry error rate even at 1000 entries, showing that interpolation reaches low error at a fraction of the storage nearest-neighbor would require. Fig.~\ref{fig:lut_resources} plots accuracy and error rate against table size for both methods on this setting, showing the rapid saturation of linear interpolation against the persistently slower convergence of nearest-neighbor.

Finally, to eliminate additional runtime overhead, the input normalization required by Eq.~\ref{algo:BernAct}: $t = (z-l_i)/(u_i-l_i)$ is fused into the preceding linear layer's weights offline:
\begin{align}
W_\mathrm{fused}[i][j] = W[i][j] / (u_i{-}l_i) \\
\beta_\mathrm{fused}[i] = (\beta[i]{-}l_i)/(u_i{-}l_i)
\end{align}
The linear layer output then arrives already normalized to $[0,1]$, so the activation requires only a clamp and an index computation, improving hardware efficiency. This fusion is only possible because Bernstein activations use fixed, learned input bounds $l_i$ and $u_i$ that do not change at runtime.

% ============================================================
\subsection{Symbolic Rule Extraction}
\label{sec:rules}
% ============================================================

The second deployment path in \sysname begins with symbolic rule extraction, as shown in Fig.~\ref{fig:framework_pipeline}. We propose an activation-geometry-based pipeline converting the trained student BNN into a compact, interpretable rule set.

\textbf{Coefficient-driven structure and analytic regime partitioning:} The coefficients $\mathbf{c}$ fully determine the shape of Bernstein activation over the normalized domain. The derivative preserves this structure through differences between adjacent coefficients:
\begin{equation}\label{eq:derivativeofBNN}
  \sigma'(z) = \frac{n}{u-l}
  \sum_{k=0}^{n-1}(c_{k+1} - c_k)\,
  b_{n-1,k}\!\left(\tfrac{z-l}{u-l}\right).
\end{equation}

The roots of $\sigma'(z)$ define local extrema partitioning the activation into monotonic regimes determined directly by the learned coefficients. In \sysname, these regimes encode the learned activation geometry and form the basis for symbolic rule extraction and subsequent hardware synthesis.

This formulation is specific to BNNs, whose coefficient-driven polynomial structure enables analytic extraction of multiple activation regimes from the learned function. Piecewise-linear activations such as ReLU, in contrast, define only a single breakpoint and lack the coefficient structure needed for multiple intrinsic regimes, making them less amenable to this extraction method.

\textbf{Pipeline of symbolic rule extraction:} The pipeline consists of three stages. Stage~1 extracts neuron-wise activation regimes from the Bernstein activation geometry in pre-activation space. Stage~2 composes these regimes into candidate rules and selects a compact subset via cascade greedy cover based on purity and coverage. Stage~3 optimizes the selected rules for deployment via weight vector sparsification and quantization, producing a hardware-efficient final rule set. We detail each stage below.

% ============================================================
\subsubsection{\textit{Stage 1: Geometry Extraction}} The first stage extracts neuron-wise activation regimes directly from
the Bernstein activation geometry in pre-activation space ($z$-space) as follows.
%============================================================
\begin{itemize}[noitemsep, topsep=0pt]
\item \textbf{Pre-activation mapping.}
Each neuron $i$ defines a scalar pre-activation (simplified from Eq.~\ref{eq:propagation}): 

\begin{equation}
z_i = \mathbf{w}_i^\top \mathbf{x} + \beta_i,
\label{eq:preactivation}
\end{equation}
mapped to a normalized domain $t \in [0,1]$ using learned input bounds $[l_i, u_i]$ (Section~\ref{sec:training}).

\item \textbf{Activation motif classification.}
The learned Bernstein coefficients $\mathbf{c}_i$ fully determine each activation's shape. We classify neurons into qualitative
\emph{motifs} (e.g., monotone, bump, valley) describing the overall
structure of the activation response, as shown in
Fig.~\ref{fig:activation_regimes}.

\item \textbf{Regime boundary extraction.}
We compute regime boundaries analytically from the activation
geometry. Breakpoints in $t$-space come from three sources:
(i)~a uniform grid ensuring baseline coverage,
(ii)~roots of the first derivative identifying local extrema, and
(iii)~roots of the second derivative identifying inflection points.
As Eq.~\ref{eq:derivativeofBNN} shows, these derivatives take closed
form in terms of coefficient differences, enabling analytic computation
of all breakpoints without dependence on data samples, reflecting
the learned activation shapes, as illustrated in Fig.~\ref{fig:activation_regimes}.

\begin{figure}[t]
  \centering
  \includegraphics[width=\columnwidth]{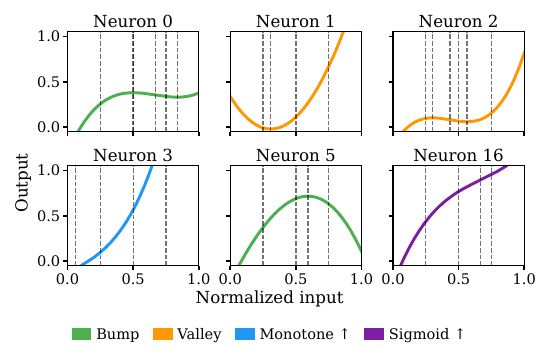}
  \caption{Bernstein activation curves with analytically derived regime breakpoints. Six representative neuron activations from a BNN ($h=64$) trained on Adult~\cite{kohavi1996scaling}, with breakpoints from derivative roots and inflection points marked.}
\label{fig:activation_regimes}
\end{figure}

\item \textbf{Input-space mapping.}
We map the breakpoints back to pre-activation space using the
fixed input bounds $[l_i, u_i]$, via
\begin{equation}
z_i = l_i + t \,(u_i - l_i),
\label{eq:t_to_z}
\end{equation}
the inverse of the normalization in Eq.~\ref{algo:BernAct}, producing
disjoint intervals that partition each neuron's response into regimes
determined entirely by its coefficients. The convex hull property of
Bernstein polynomials~\cite{farouki2012bernstein} bounds the activation
output by the range of its coefficients; restricted to subintervals,
the effective bounds tighten further, reinforcing the stability of
the resulting regime partitions.

The mapping of these regimes to input-space constraints is illustrated
step-by-step in Fig.~\ref{fig:activation_geometry_rule}. We apply this
mapping to first-layer neurons, whose pre-activations are affine in the
raw input $\mathbf{x}$ (Eq.~\ref{eq:preactivation}); a selected interval
in $t$-space defines a corresponding interval in $z$-space via
Eq.~\ref{eq:t_to_z}, which, through Eq.~\ref{eq:preactivation}, directly
induces an affine constraint in input space of the form
\begin{equation}
z_i^{\text{lo}} \le \mathbf{w}_i^\top \mathbf{x} + \beta_i \le z_i^{\text{hi}},
\label{eq:input_constraint}
\end{equation}
defining an oblique band in input space. Each condition in a rule
corresponds to one neuron and one of its activation regimes; the
conjunction of conditions across multiple neurons produces intersecting
affine bands, forming oblique polyhedral regions in the input space.
Stage~2 constructs candidate rules from these intervals and refines them
based on empirical quality.

\begin{figure*}[t]
    \centering
    \includegraphics[scale=0.6, trim={0.5 3cm 0.5cm 3cm}, clip]{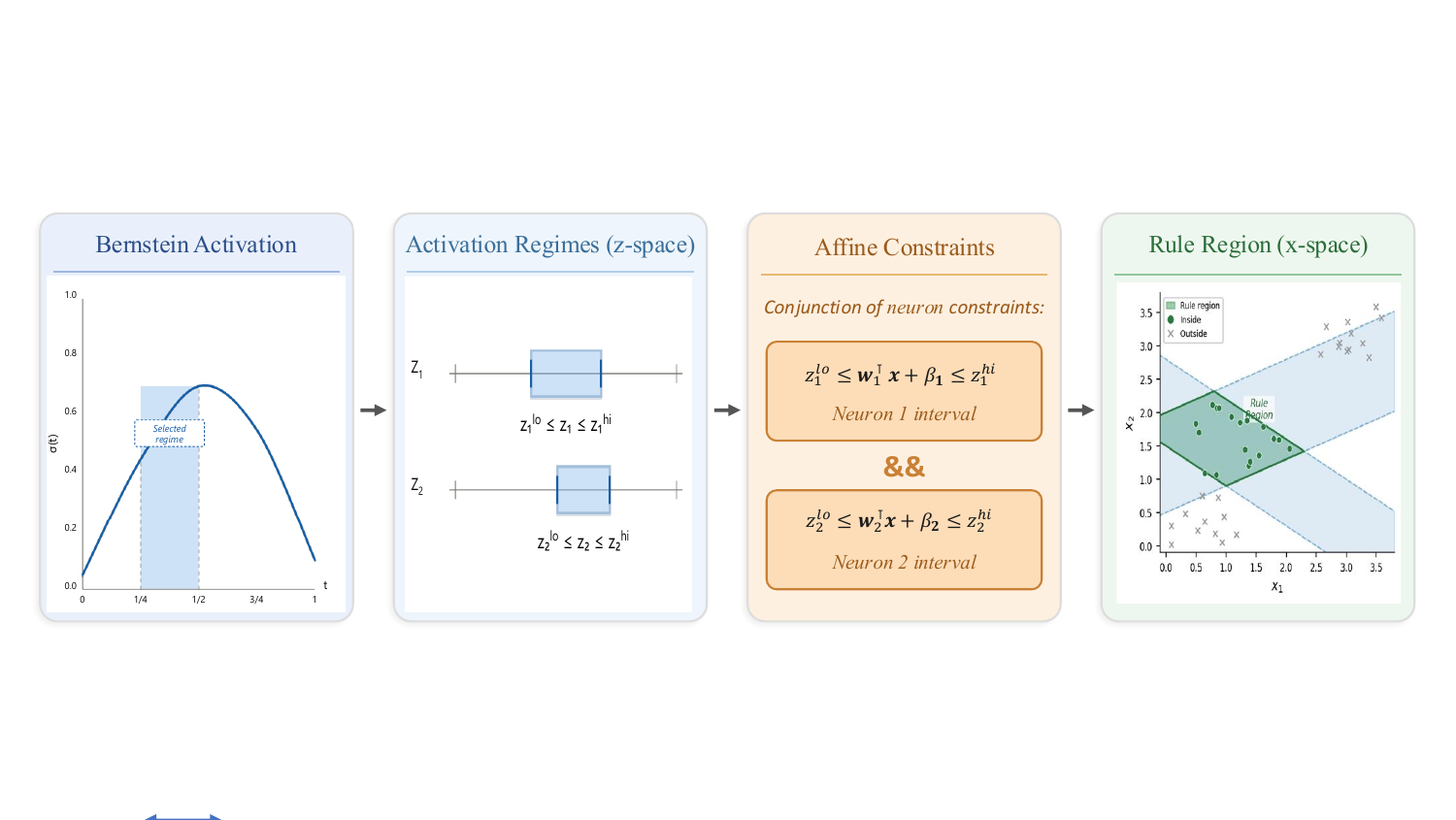}
    \caption{Activation-geometry-based rule formation. A selected regime in
normalized activation space ($t$) maps through $z$-space to an affine
constraint in input space, forming interpretable oblique regions.}
\label{fig:activation_geometry_rule}
\end{figure*}
 \end{itemize}
% ============================================================
% ============================================================
\subsubsection{\textit{Stage 2: Candidate Rule Generation and Selection}}
Given the regime partitions from Stage~1, we define rules as conjunctions of affine constraints of the
form in Eq.~\ref{eq:input_constraint}, one per neuron regime, assigning each rule the majority predicted class of its covered samples.

\begin{itemize}[noitemsep, topsep=0pt]
\item \textbf{Rule evaluation.}
We evaluate each candidate rule $r$ on training data by computing
the set of samples satisfying all its conditions, from which we compute: (i)~\emph{purity}, the fraction of covered samples whose prediction matches the rule's label, and (ii)~\emph{coverage}, the number of samples matched. We discard rules that do not
meet minimum coverage and purity thresholds; both are hyperparameters of the method.

\item \textbf{Candidate generation.}
We construct rules progressively by combining regimes across neurons, beginning with single-condition rules evaluated for all neurons and their regimes. Rules not meeting the desired purity become \emph{impure seeds}, iteratively expanded by adding conditions from additional neurons up to a predefined depth limit.

At each expansion step, we evaluate candidate extensions formed by adding one condition to each impure parent rule, retaining two: (i)~the highest-coverage extension meeting
the purity criterion, and (ii)~the highest-coverage extension that does not, serving as the seed for further expansion. This retains at
most one pure and one impure extension per parent rule, bounding candidate growth as neurons and regimes increase while preserving the highest-coverage candidates.

\item \textbf{Cascade greedy selection.}
From the candidate pool, we select a compact rule set via cascade greedy cover. The cascade proceeds over decreasing
purity thresholds with increasing minimum coverage requirements,
ensuring lower-purity rules are selected only if they add substantial new coverage.

Within each stage, we select rules iteratively using a scoring
function prioritizing new-sample coverage while penalizing redundancy
and conflicts:
\begin{equation}
\text{score}(r) = \text{gain}(r) 
- \alpha_{\text{sc}} \cdot \text{same\textrm{-}cover}(r)
- \alpha_{\text{conf}} \cdot \text{conf}(r).
\label{eq:rule_penalties}
\end{equation}
Here, $\text{gain}(r)$ is the number of samples covered by $r$ not yet covered by the current rule set. $\text{same\textrm{-}cover}(r)$ counts overlap with selected rules of the same label, while $\text{conf}(r)$ counts overlap with rules of different labels. The weights $\alpha_{\text{sc}}$ and $\alpha_{\text{conf}}$ control the corresponding penalties; their effect on coverage and accuracy is analyzed in Section~\ref{sec:ablation}.

At each step, we select the rule with the highest score and sufficient new coverage, updating coverage accordingly, until no further rules can be added, after which the algorithm
proceeds to the next lower purity stage until the minimum purity threshold is reached.

This cascade yields a compact rule set capturing high-confidence regions first, then expanding coverage while controlling redundancy and conflicts, providing a geometry-aligned approximation of the model's decision boundary (Fig.~\ref{fig:rule_partition}).

\begin{figure}[t]
  \centering
  \includegraphics[width=0.7\columnwidth]{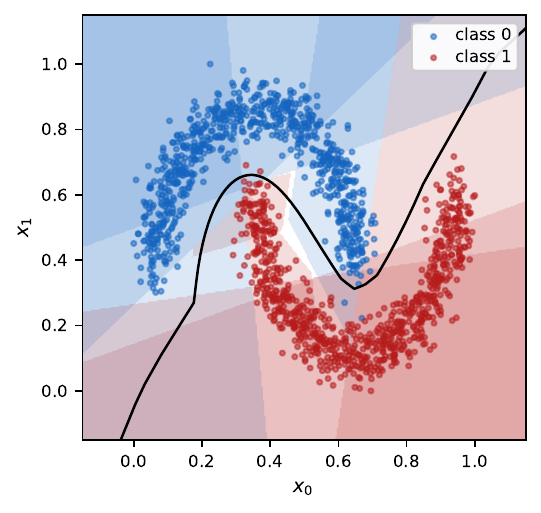}
  \caption{Activation-geometry rule regions on the Two Moons dataset~\cite{pedregosa2011scikit}. Rules align with the curved decision boundary via Bernstein-derived breakpoints, enabling compact partitioning.}
  \label{fig:rule_partition}
\end{figure}

\end{itemize}

% ============================================================
\subsubsection{\textit{Stage 3: Rule Optimization}}
% ============================================================
To enable efficient inference on resource-constrained hardware,
we optimize the extracted rules for reduced computation and memory usage as follows.

\begin{itemize}[noitemsep, topsep=0pt]
\item \textbf{Rule sparsification.}
Each rule condition takes the affine form of Eq.~\ref{eq:input_constraint}.
In practice, a small subset of large-magnitude weights often dominates
the output, indicating only a few input features contribute
significantly. We therefore retain only the top-$k$ entries of each
weight vector, zeroing the rest, reducing both
storage and inference cost to $k$ multiplications per condition. $k$ is a hyperparameter; its effect on memory
footprint and accuracy is ablated in Section~\ref{sec:ablation}.

\item \textbf{Integer quantization.}
We quantize rule parameters to integer precision for direct
deployment on integer-only hardware. Weight vectors use symmetric
per-vector INT8 quantization:
\[
\mathbf{w}_q = \mathrm{round}\!\left(\frac{\mathbf{w}}{s}\right), 
\quad s = \frac{\max_j |w_j|}{127},
\]
dequantized as $\hat{\mathbf{w}} = s \cdot \mathbf{w}_q$; thresholds and biases are quantized to fixed-point precision.

Unlike multi-layer neural networks, quantization here introduces no
cross-layer error accumulation, since each rule evaluates as a
single affine operation on the original input. As a result, INT8
quantization introduces negligible degradation in rule accuracy, with an
average accuracy drop of $0.002\%$ across all tested datasets.

The final rule set consists of sparse, quantized constraints evaluable with integer and fixed-point arithmetic, enabling hardware-efficient inference.
\end{itemize}
% ============================================================
\subsubsection{\textit{Inference}}
% ============================================================
The final stage performs rule inference as follows.

\begin{itemize}[noitemsep, topsep=0pt]
\item \textbf{Conflict resolution.}
Due to overlapping regimes, a sample may satisfy multiple rules with
different labels. We resolve conflicts by selecting the matching
rule with the highest training purity:
\begin{equation}
\hat{y} = \arg\max_{r \in \mathcal{R}(\mathbf{x})} \text{purity}(r),
\end{equation}
where $\mathcal{R}(\mathbf{x})$ is the set of rules whose conditions are all satisfied by sample $\mathbf{x}$, and $\text{purity}(r)$ is the purity of rule $r$ measured on the training set (Stage~2).

\item \textbf{Fallback.}
Inputs not matched by any rule fall through to a Classification
and Regression Tree (CART) trained on the uncovered training samples for
which no rule fires~\cite{breiman1984cart}. The CART partitions the
feature space via threshold comparisons on raw features, requiring no
multiplications. We keep the tree shallow (depth~4) to preserve
interpretability and minimize hardware cost, quantizing
its split thresholds to fixed-point \texttt{fix\textless16,8\textgreater}
before synthesis. We compare CART against alternative fallback strategies
(linear regression, a small BNN trained on uncovered samples, and the
full underlying BNN) in Section~\ref{sec:ablation}
(Table~\ref{tab:fallback_ablation}).

\end{itemize}

\subsubsection{\textit{Rule Synthesis}}
We synthesize each rule set into hardware operating in three
pipelined phases. Phase~1 computes per-condition linear projections as
sparse INT8 dot products: each condition stores only non-zero quantized
weights alongside their feature indices in on-chip ROM, reducing multiply
operations and weight storage relative to dense evaluation. We fully
unroll and pipeline the dot product, dequantizing the result by a
per-condition scale factor. Phase~2 evaluates each rule by checking
whether its condition projections fall within learned band boundaries
(lower/upper thresholds); the highest-purity firing rule determines the
predicted label. Phase~3 routes samples not matched by any rule to the
CART fallback. All rule parameters reside as compile-time ROM constants,
requiring no external memory access.

\section{Results}
\label{sec:results}
% ============================================================
\subsection{Experimental Setup}
\label{sec:setup}
% ============================================================

\noindent \textbf{$\bullet$ Datasets and Preprocessing.}
We evaluate \sysname on tabular benchmarks from prior
work~\cite{gorishniy2021}, and a language classification
task for the transformer generalization study.
In particular, \textbf{HIGGS-Small}~\cite{baldi2014higgs,vanschoren2014openml}
(98,049 samples, 28 features, binary),
\textbf{Covertype}~\cite{blackard1999comparative}
(581,012 samples, 54 features, 7 classes), and
\textbf{Adult Census}~\cite{kohavi1996scaling} (48,842 samples, 14 mixed features, binary). These datasets use standard preprocessing and stratified splits following Gorishniy et al.~\cite{gorishniy2021}.
\textbf{MAGIC Gamma Telescope}~\cite{magic_gamma_telescope_159}
(19,020 samples, 10 features, binary) is used for rule extraction
benchmarking and rule certification.
\textbf{ACS Income}~\cite{ding2021retiring}, a published covariate-shift
benchmark, evaluates rule-system robustness under geographic and temporal
distribution shift.
\textbf{SST-2}~\cite{socher2013recursive} evaluates Bernstein
activations in transformer FFN sublayers, with
TinyBERT4~\cite{jiao2020tinybert} as the target architecture.
Unless otherwise noted, all experiments use 5-fold cross-validation; test data are transformed
using preprocessors fitted on the training split. Rule extraction results (Sections~\ref{sec:results_symbolic}, IV-F) use single representative models per configuration.

\noindent \textbf{$\bullet$ Teacher and Student Models.}
Teacher models replicate the exact MLP configurations of
Gorishniy et al.~\cite{gorishniy2021}, achieving accuracy within $0.5$ percentage points (pp) of reported baselines.
Students are trained via KD~\cite{hinton2015distilling}
using identical architectures with either Bernstein activations
(degree~3 for MLPs; degree~15 for transformer FFN sublayers) or ReLU. Training uses AdamW with learning rate $6\times10^{-3}$, weight
decay $1\times10^{-4}$, and cosine annealing; BNNs additionally include an out-of-bounds penalty ($\lambda_{\mathrm{ob}}=10^{-2}$, Section~\ref{sec:training}). Distillation parameters are swept per
dataset with temperature $T \in \{2,4\}$ and $\alpha \in [0,0.85]$, and
the best-performing configuration is reported. Student architectures are
selected by sweeping width and depth under hardware constraints. We denote architectures as $\{d_{\text{in}}, h_1, \ldots, h_m, d_{\text{out}}\}$; where a single hidden layer is varied, we write its width as a scalar~$h$.

\noindent \textbf{$\bullet$ Hardware Setup.}
Training uses an NVIDIA Tesla V100 (16\,GB). Designs are synthesized with
Vitis HLS 2024.1 and Vivado 2024.1 and evaluated on an AMD Xilinx KV260
FPGA at 200\,MHz. We also evaluate on a Spartan-7 XC7S15 FPGA to
demonstrate low-power edge deployment.

\noindent \textbf{$\bullet$ Evaluation.}
We evaluate three components: (1) neural compression, comparing Bernstein
and ReLU students under identical constraints; (2) symbolic rule
extraction; and (3) the full \sysname pipeline against
a W8A8 (8-bit weight, 8-bit activation) quantization-aware training (QAT) teacher baseline~\cite{Jacob_2018_CVPR}. We further conduct three additional studies: low-power deployment on
the Spartan-7 XC7S15, ablation on the rule extraction hyperparameters ($k$, $\alpha_{\text{sc}}$, $\alpha_{\text{conf}}$) and fallback strategies, and rule robustness evaluation covering input-noise certification and distribution-shift assessment on benchmarks.

\noindent \textbf{$\bullet$ Metrics.} We report test
accuracy and cross-entropy loss. For hardware deployment, we report latency, DSPs, BRAM, LUTs, and flip-flops (FFs). For rule extraction, we additionally report rule count, average conditions per rule, and coverage.

% ============================================================
\subsection{Student BNN Compression and Synthesis}
\label{sec:results_kd}
% ============================================================

Table~\ref{tab:combined_compression} reports accuracy and synthesis
metrics for BNNs and equivalent ReLU student networks across three datasets.
Teacher models follow the MLP baselines of Gorishniy et
al.~\cite{gorishniy2021}: $\{28,231,121,2\}$ on HIGGS-Small, $\{54,1024,1024,512,7\}$ on Covertype, and $\{14,503,503,503,111,2\}$ on Adult.
% ─── Main compression + synthesis table ──────────────────────────────────────
\begin{table*}[t]
\centering
\caption{Accuracy and synthesis metrics on AMD KV260 (5-fold CV). Latency in clock
cycles. $\Delta$Acc\,/\,$\Delta$CE: BNN minus ReLU.}
{%
\scriptsize
\setlength{\tabcolsep}{7pt}
\renewcommand{\arraystretch}{1.3}
\begin{tabular}{c|c|c|c|c|c|c|c|c|c}
\hline
Dataset & Arch. & Act. & Acc. (\%) & CE Loss & $\Delta$Acc\,/\,$\Delta$CE
  & Latency & DSPs & BRAMs & LUTs \\
\hline
\multirow{6}{*}{\rotatebox{90}{HIGGS-Small}}
& \multirow{2}{*}{$\{28,16,2\}$}
  & ReLU & 70.98 $\pm$ 0.20 & 0.553
  & \multirow{2}{*}{$+0.94$ / $-0.013$}
  & \textbf{85} & 108 & \textbf{0} & \textbf{5397} \\
&& Bern  & \textbf{71.92} $\pm$ 0.26 & \textbf{0.540}
  && 87 & \textbf{106} & 2 & 5682 \\
\cline{2-10}
& \multirow{2}{*}{$\{28,16,8,2\}$}
  & ReLU & 71.84 $\pm$ 0.24 & 0.540
  & \multirow{2}{*}{$+0.48$ / $-0.008$}
  & \textbf{121} & 156 & \textbf{0} & \textbf{6852} \\
&& Bern  & \textbf{72.32} $\pm$ 0.18 & \textbf{0.532}
  && 125 & \textbf{135} & 3 & 7294 \\
\cline{2-10}
& \multirow{2}{*}{$\{28,128,2\}$}
  & ReLU & 72.08 $\pm$ 0.27 & 0.524
  & \multirow{2}{*}{$+0.45$ / $-0.012$}
  & \textbf{337} & 108 & \textbf{34} & \textbf{5114} \\
&& Bern  & \textbf{72.53} $\pm$ 0.18 & \textbf{0.512}
  && 339 & \textbf{103} & 44 & 5410 \\
\hline
\multirow{6}{*}{\rotatebox{90}{Covertype}}
& \multirow{2}{*}{$\{54,64,32,7\}$}
  & ReLU & 88.97 $\pm$ 0.82 & 0.292
  & \multirow{2}{*}{$+2.12$ / $-0.069$}
  & \textbf{960} & 360 & 113 & \textbf{20213} \\
&& Bern  & \textbf{91.09} $\pm$ 0.27 & \textbf{0.223}
  && 963 & \textbf{326} & \textbf{105} & 21815 \\
\cline{2-10}
& \multirow{2}{*}{$\{54,128,64,7\}$}
  & ReLU & 93.53 $\pm$ 0.18 & 0.138
  & \multirow{2}{*}{$+1.52$ / $-0.060$}
  & \textbf{2477} & 360 & 137 & \textbf{22207} \\
&& Bern  & \textbf{95.05} $\pm$ 0.05 & \textbf{0.078}
  && 2481 & \textbf{326} & \textbf{136} & 23824 \\
\cline{2-10}
& \multirow{2}{*}{$\{54,256,128,7\}$}
  & ReLU & 95.62 $\pm$ 0.12 & 0.057
  & \multirow{2}{*}{$+0.77$ / $-0.029$}
  & \textbf{7429} & 360 & \textbf{169} & \textbf{22432} \\
&& Bern  & \textbf{96.39} $\pm$ 0.02 & \textbf{0.028}
  && 7433 & \textbf{326} & 179 & 23063 \\
\hline
\multirow{6}{*}{\rotatebox{90}{Adult}}
& \multirow{2}{*}{$\{14,16,2\}$}
  & ReLU & 84.06 $\pm$ 0.39 & 0.324
  & \multirow{2}{*}{$+0.54$ / $-0.010$}
  & \textbf{68} & 69 & \textbf{0} & \textbf{3823} \\
&& Bern  & \textbf{84.60} $\pm$ 0.13 & \textbf{0.314}
  && 70 & \textbf{69} & 2 & 4113 \\
\cline{2-10}
& \multirow{2}{*}{$\{14,32,16,2\}$}
  & ReLU & 84.48 $\pm$ 0.12 & 0.315
  & \multirow{2}{*}{$+0.44$ / $-0.009$}
  & \textbf{161} & 163 & \textbf{1} & \textbf{7739} \\
&& Bern  & \textbf{84.92} $\pm$ 0.13 & \textbf{0.306}
  && 165 & \textbf{136} & 6 & 8132 \\
\cline{2-10}
& \multirow{2}{*}{$\{14,128,2\}$}
  & ReLU & 84.57 $\pm$ 0.22 & \textbf{0.310}
  & \multirow{2}{*}{$+0.25$ / $+0.001$}
  & \textbf{321} & 67 & \textbf{20} & \textbf{3595} \\
&& Bern  & \textbf{84.82} $\pm$ 0.14 & 0.311
  && 323 & \textbf{61} & 30 & 3892 \\
\hline
\end{tabular}%
}
\label{tab:combined_compression}
\end{table*}

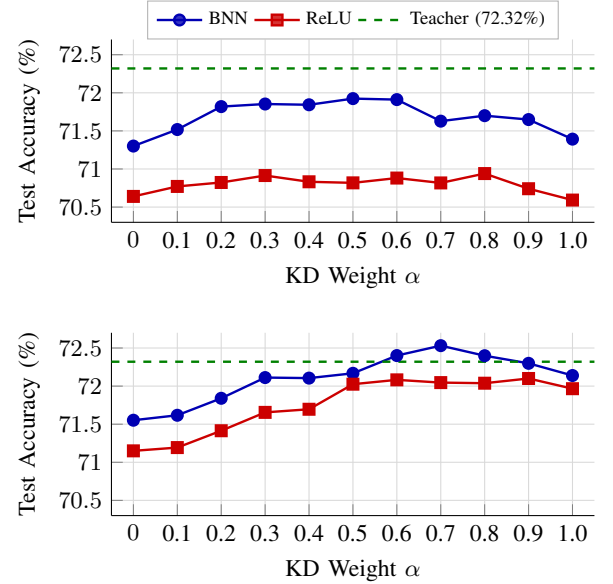
\begin{figure}[t]
  \centering
  \subfloat{\begin{tikzpicture}
\begin{axis}[
    width=0.9\columnwidth,
    height=4cm,
    title style={font=\small, yshift=-2pt},
    xlabel={KD Weight $\alpha$},
    ylabel={Test Accuracy (\%)},
    axis y line*=left,
    axis x line*=bottom,
    xtick={0.0,0.1,0.2,0.3,0.4,0.5,0.6,0.7,0.8,0.9,1.0},
    xticklabels={$0$,0.1,0.2,0.3,0.4,0.5,0.6,0.7,0.8,0.9,1.0},
    xmin=-0.05, xmax=1.05,
    ymin=70.3, ymax=72.7,
    ytick={70.0,70.5,71.0,71.5,72.0,72.5,73.0},
    grid=both,
    grid style={line width=0.4pt, draw=gray!30},
    tick label style={font=\small},
    label style={font=\small},
    legend style={
        at={(0.5,1)},
        anchor=south,
        font=\scriptsize,
        draw=gray!60,
        fill=white,
        fill opacity=0.8,
        text opacity=1,
        legend columns=3
    },
    legend cell align={center},
]

%% Bernstein (degree 3)
\addplot[
    color=blue!70!black,
    solid,
    mark=*,
    mark size=2pt,
    line width=0.9pt,
] coordinates {
    (0.0, 71.301)
    (0.1, 71.518)
    (0.2, 71.819)
    (0.3, 71.853)
    (0.4, 71.843)
    (0.5, 71.924)
    (0.6, 71.911)
    (0.7, 71.628)
    (0.8, 71.700)
    (0.9, 71.649)
    (1.0, 71.392)
};
\addlegendentry{BNN}

%% ReLU
\addplot[
    color=red!80!black,
    solid,
    mark=square*,
    mark size=2pt,
    line width=0.9pt,
] coordinates {
    (0.0, 70.640)
    (0.1, 70.771)
    (0.2, 70.823)
    (0.3, 70.914)
    (0.4, 70.832)
    (0.5, 70.818)
    (0.6, 70.881)
    (0.7, 70.817)
    (0.8, 70.940)
    (0.9, 70.741)
    (1.0, 70.592)
};
\addlegendentry{ReLU}

%% Teacher baseline
\addplot[
    color=green!50!black,
    dashed,
    line width=0.9pt,
    forget plot,
] coordinates {(-0.05, 72.32) (1.05, 72.32)};

\addlegendimage{color=green!50!black, dashed, line width=0.9pt}
\addlegendentry{Teacher (72.32\%)}

\end{axis}
\end{tikzpicture}\label{fig:kd_small}}
  \hfill
  \subfloat{\begin{tikzpicture}
\begin{axis}[
    width=0.9\columnwidth,
    height=4cm,
    title style={font=\small, yshift=-2pt},
    xlabel={KD Weight $\alpha$},
    ylabel={Test Accuracy (\%)},
    axis y line*=left,
    axis x line*=bottom,
    xtick={0.0,0.1,0.2,0.3,0.4,0.5,0.6,0.7,0.8,0.9,1.0},
    xticklabels={$0$,0.1,0.2,0.3,0.4,0.5,0.6,0.7,0.8,0.9,1.0},
    xmin=-0.05, xmax=1.05,
    ymin=70.3, ymax=72.7,
    ytick={70.5,71.0,71.5,72.0,72.5,73.0},
    grid=both,
    grid style={line width=0.4pt, draw=gray!30},
    tick label style={font=\small},
      label style={font=\small},
    legend style={
        at={(0.5,1)},
        anchor=south,
        font=\scriptsize,
        draw=gray!60,
        fill=white,
        fill opacity=0.8,
        text opacity=1,
        legend columns=3
    },
    legend cell align={center},
]

%% Bernstein (degree 3)
\addplot[
    color=blue!70!black,
    solid,
    mark=*,
    mark size=2pt,
    line width=0.9pt,
] coordinates {
    (0.0, 71.552)
    (0.1, 71.617)
    (0.2, 71.840)
    (0.3, 72.112)
    (0.4, 72.105)
    (0.5, 72.168)
    (0.6, 72.401)
    (0.7, 72.531)
    (0.8, 72.399)
    (0.9, 72.299)
    (1.0, 72.139)
};

%% ReLU
\addplot[
    color=red!80!black,
    solid,
    mark=square*,
    mark size=2pt,
    line width=0.9pt,
] coordinates {
    (0.0, 71.150)
    (0.1, 71.193)
    (0.2, 71.414)
    (0.3, 71.655)
    (0.4, 71.696)
    (0.5, 72.025)
    (0.6, 72.082)
    (0.7, 72.046)
    (0.8, 72.038)
    (0.9, 72.100)
    (1.0, 71.966)
};

%% Teacher baseline
\addplot[
    color=green!50!black,
    dashed,
    line width=0.9pt,
    forget plot,
] coordinates {(-0.05, 72.32) (1.05, 72.32)};

\end{axis}
\end{tikzpicture}\label{fig:kd_large}}
  \caption{%
Test accuracy vs.\ KD weight $\alpha$ ($T{=}2$) on HIGGS-Small for students with single layer $h=16$~(top) and $h=128$~(bottom). Dashed line: teacher baseline ($72.32\%$).
  } 
  \label{fig:kd_sensitivity_teacher}
\end{figure}

\noindent \textbf{$\bullet$ Accuracy Under Compression.}
BNNs consistently outperform their ReLU counterparts across all
architecture--dataset combinations (Table~\ref{tab:combined_compression}), with
accuracy gains of $+0.25$ to $+2.12$\,pp and lower cross-entropy loss in 8 of 9
cases. The gap is largest under strong compression, reaching $+2.12$\,pp on
Covertype~$\{54,64,32,7\}$, and narrows as model capacity increases, consistent
with the approximation behavior of Bernstein-based networks~\cite{albool2026deepbern}.
BNNs also approach or exceed teacher accuracy at smaller architectures than
ReLU on every dataset. Gains on Adult and HIGGS-Small are modest but consistent
across all folds; since inter-model differences below one percentage point are
typical for state-of-the-art tabular models~\cite{gorishniy2021}, these
improvements reflect genuine learning efficiency rather than incidental variation.

\noindent \textbf{$\bullet$ Alignment with Teacher under KD.}
Fig.~\ref{fig:kd_sensitivity_teacher} shows test accuracy vs.\ $\alpha$ on
HIGGS-Small for a small student~($h=16$) (top) and a large student~($h=128$)
(bottom). For the small student, the BNN holds a stable ${\sim}1$\,pp lead
over ReLU across all $\alpha$ values, with neither reaching the teacher; width
is the binding constraint, and the Bernstein advantage persists even without
distillation ($\alpha{=}0$). For the larger student, both activations improve
with $\alpha$, but the BNN crosses the teacher near $\alpha{=}0.7$ and peaks
at $72.5\%$, while ReLU approaches but does not reach it. Thus at low capacity
the Bernstein advantage is intrinsic and KD-invariant, while at higher capacity
the BNN additionally benefits more from stronger distillation.

\noindent \textbf{$\bullet$ Hardware Overhead.}
Bernstein activations impose modest, predictable hardware overhead relative
to ReLU at matched architectures (Table~\ref{tab:combined_compression}).
Latency overhead never exceeds 4 clock cycles, since each per-neuron Bernstein
evaluation reduces to a single LUT lookup followed by linear interpolation.
Bernstein uses fewer or equal DSPs than ReLU at every architecture, with
savings reaching 27 on Adult~$\{14,32,16,2\}$: the HLS compiler prunes unused
upper bits from the activation output, so each downstream multiply fits in
2~DSPs instead of 3, whereas unbounded ReLU preserves the full fixed-point
\texttt{fix<32,16>} range and requires 3~DSPs per multiply. LUT counts are
higher for Bernstein across all configurations ($2.8$--$8.3\%$ overhead),
reflecting the synthesized per-neuron lookup tables; BRAM overhead is small
and architecture-dependent, governed by the interaction between per-neuron
table size and BRAM tile granularity.

\noindent \textbf{$\bullet$ Accuracy Under Matched Hardware Budgets.}
Table~\ref{tab:covertype_hw_comparison} selects the best Bernstein and ReLU
models on Covertype within five hardware budget tiers, confirming Bernstein
outperforms ReLU under identical tight constraints at every operating point.
Gains range from $+0.77$\,pp at the 7500-cycle tier to $+2.12$\,pp at
1000 cycles, ruling out parameter count and architecture size as confounds.

\begin{table}[t]
\centering
\caption{Covertype accuracy under matched latency and BRAM budgets.
Parentheses: actual synthesis values (cycles, BRAMs).} 
\label{tab:covertype_hw_comparison}
\setlength{\tabcolsep}{2.5pt}
\scriptsize
\renewcommand{\arraystretch}{1.3}
\resizebox{\columnwidth}{!}{%
\begin{tabular}{ccrlrlr}
\toprule
Latency & BRAMs & \multicolumn{2}{c}{Bernstein} &
  \multicolumn{2}{c}{ReLU} & $\Delta$Acc \\
(cyc) & & Acc (\%) & (cyc,\ BRAM) &
  Acc (\%) & (cyc,\ BRAM) & (pp) \\
\midrule
  ${\leq}200$ & ${\leq}30$  & \textbf{76.50} $\pm$ 0.44 & (189,\ 26)
    & 74.93 $\pm$ 0.08 & (188,\ 25) & $+1.57$ \\
  ${\leq}400$ & ${\leq}55$  & \textbf{82.98} $\pm$ 0.16 & (397,\ 52)
    & 81.87 $\pm$ 0.37 & (396,\ 49) & $+1.11$ \\
 ${\leq}1000$ & ${\leq}120$ & \textbf{91.09} $\pm$ 0.27 & (963,\ 105)
    & 88.97 $\pm$ 0.82 & (960,\ 113) & $+2.12$ \\
 ${\leq}2500$ & ${\leq}140$ & \textbf{95.05} $\pm$ 0.05 & (2481,\ 136)
    & 93.53 $\pm$ 0.18 & (2477,\ 137) & $+1.52$ \\
 ${\leq}7500$ & ${\leq}180$ & \textbf{96.39} $\pm$ 0.02 & (7433,\ 179)
    & 95.62 $\pm$ 0.12 & (7429,\ 169) & $+0.77$ \\
\bottomrule
\end{tabular}
}
\end{table}

\begin{figure}[t]
  \centering
  \begin{tikzpicture}
\begin{axis}[
  width            = \columnwidth,
  height           = 4cm,
  ybar,
  bar width        = 4pt,
  enlarge x limits = 0.18,
  ylabel           = {Resource Util. (\%)},
  ylabel style     = {font=\small},
  xtick            = {1,2,3,4},
  xticklabels      = {BRAM, DSP, FF, LUT},
  xticklabel style = {font=\small},
  yticklabel style = {font=\footnotesize},
  ytick align      = inside,
  xtick align      = inside,
  ymajorgrids      = true,
  grid style       = {dashed, gray!40},
  axis lines*      = left,
  clip             = false,
  legend columns   = 5,
  legend image post style={xscale=0.7},
  legend style     = {
    at         = {(0.5,1.05)},
    anchor     = south,
    font       = \footnotesize,
    draw       = gray!50,
    fill       = white,
    column sep = 0.1em,
    inner sep  = 3pt,
  },
]

% Naive Bernstein evaluation
\addplot[
  ybar,
  forget plot,
  bar shift=-21pt,
  fill=naiveA,
  draw=naiveA,
  line width=0.3pt
] coordinates {
  (1,11.1111) (2,6.8109) (3,1.6487) (4,3.5126)
};

\addplot[
  ybar,
  forget plot,
  bar shift=-16.5pt,
  fill=naiveB,
  draw=naiveB!60,
  line width=0.3pt,
  postaction={
    pattern=north east lines,
    pattern color=naiveA
  }
] coordinates {
  (1,11.8056) (2,8.0128) (3,1.7491) (4,3.7739)
};

\addplot[
  ybar,
  forget plot,
  bar shift=-12pt,
  fill=naiveC,
  draw=naiveC!60,
  line width=0.3pt,
  postaction={
    pattern=crosshatch,
    pattern color=naiveA
  }
] coordinates {
  (1,12.5000) (2,9.4551) (3,1.9873) (4,4.1539)
};

\addplot[
  ybar,
  forget plot,
  bar shift=-7.5pt,
  fill=naiveD,
  draw=naiveD!60,
  line width=0.3pt,
  postaction={
    pattern=dots,
    pattern color=naiveA
  }
] coordinates {
  (1,13.1944) (2,10.7372) (3,2.2733) (4,4.4962)
};

\addplot[
  ybar,
  forget plot,
  bar shift=-3pt,
  fill=naiveE,
  draw=naiveE!60,
  line width=0.3pt,
  postaction={
    pattern=horizontal lines,
    pattern color=naiveA
  }
] coordinates {
  (1,13.8889) (2,12.0192) (3,2.4526) (4,4.8087)
};

% LUT-based evaluation
\addplot[
  ybar,
  forget plot,
  bar shift=3pt,
  fill=lutA,
  draw=lutA!,
  line width=0.3pt
] coordinates {
  (1,10.4167) (2,4.8878) (3,1.5838) (4,3.3231)
};

\addplot[
  ybar,
  forget plot,
  bar shift=7.5pt,
  fill=lutB,
  draw=lutB!60,
  line width=0.3pt,
  postaction={
    pattern=north east lines,
    pattern color=lutA!
  }
] coordinates {
  (1,10.4167) (2,4.8878) (3,1.5826) (4,3.3214)
};

\addplot[
  ybar,
  forget plot,
  bar shift=12pt,
  fill=lutC,
  draw=lutC!60,
  line width=0.3pt,
  postaction={
    pattern=crosshatch,
    pattern color=lutA!
  }
] coordinates {
  (1,10.4167) (2,4.8878) (3,1.5800) (4,3.3205)
};

\addplot[
  ybar,
  forget plot,
  bar shift=16.5pt,
  fill=lutD,
  draw=lutD!60,
  line width=0.3pt,
  postaction={
    pattern=dots,
    pattern color=lutA!
  }
] coordinates {
  (1,10.4167) (2,4.8878) (3,1.5770) (4,3.3308)
};

\addplot[
  ybar,
  forget plot,
  bar shift=21pt,
  fill=lutE,
  draw=lutE!60,
  line width=0.3pt,
  postaction={
    pattern=horizontal lines,
    pattern color=lutA!
  }
] coordinates {
  (1,10.4167) (2,4.8878) (3,1.5796) (4,3.3325)
};

% Naive legend row
\addlegendimage{
  area legend,
  fill=naiveA,
  draw=naiveA
}
\addlegendentry{Naive $n{=}3$}

\addlegendimage{
  area legend,
  fill=naiveB,
  draw=naiveB!60,
  postaction={
    pattern=north east lines,
    pattern color=naiveA
  }
}
\addlegendentry{$n{=}5$}

\addlegendimage{
  area legend,
  fill=naiveC,
  draw=naiveC!60,
  postaction={
    pattern=crosshatch,
    pattern color=naiveA
  }
}
\addlegendentry{$n{=}7$}

\addlegendimage{
  area legend,
  fill=naiveD,
  draw=naiveD!60,
  postaction={
    pattern=dots,
    pattern color=naiveA
  }
}
\addlegendentry{$n{=}9$}

\addlegendimage{
  area legend,
  fill=naiveE,
  draw=naiveE!60,
  postaction={
    pattern=horizontal lines,
    pattern color=naiveA
  }
}
\addlegendentry{$n{=}11$}

% LUT legend row
\addlegendimage{
  area legend,
  fill=lutA,
  draw=lutA!
}
\addlegendentry{LUT $n{=}3$}

\addlegendimage{
  area legend,
  fill=lutB,
  draw=lutB!60,
  postaction={
    pattern=north east lines,
    pattern color=lutA!
  }
}
\addlegendentry{$n{=}5$}

\addlegendimage{
  area legend,
  fill=lutC,
  draw=lutC!60,
  postaction={
    pattern=crosshatch,
    pattern color=lutA!
  }
}
\addlegendentry{$n{=}7$}

\addlegendimage{
  area legend,
  fill=lutD,
  draw=lutD!60,
  postaction={
    pattern=dots,
    pattern color=lutA!
  }
}
\addlegendentry{$n{=}9$}

\addlegendimage{
  area legend,
  fill=lutE,
  draw=lutE!60,
  postaction={
    pattern=horizontal lines,
    pattern color=lutA!
  }
}
\addlegendentry{$n{=}11$}

\end{axis}
\end{tikzpicture}
  \caption{FPGA resource utilization on Adult ($h=128$) for
  naive Bernstein evaluation and LUT-based inference across polynomial degrees
  $n \in \{3,5,7,9,11\}$.}
  \label{fig:hw_resources}
\end{figure}
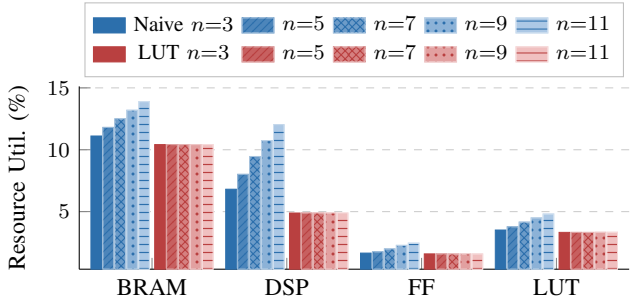

\noindent \textbf{$\bullet$ Polynomial Degree and Hardware.}
\label{sec:degree_vs_hw}
Fig.~\ref{fig:hw_resources} compares FPGA resource utilization for naive
Bernstein evaluation, which directly computes Eq.~\ref{algo:BernAct}, against
our LUT-based implementation across degrees $n \in \{3,5,7,9,11\}$.
Naive evaluation scales with degree because each additional basis function
requires more multiply-accumulate hardware, causing DSP and LUT usage to grow
monotonically. In contrast, our implementation has constant resource cost:
every activation maps to the same fixed-size lookup table at inference time.
This decouples degree selection from hardware budget, allowing $n$ to increase
for harder approximation tasks~\cite{albool2026deepbern} without additional
FPGA overhead.

% ============================================================
\subsection{Symbolic Extraction Results}
\label{sec:results_symbolic}

\noindent \textbf{$\bullet$ Rule Extraction on Adult.}
Rules are extracted from BNNs trained with
$T{=}2$ and $\alpha{=}0.5$ on Adult. The extraction pipeline
(Section~\ref{sec:rules}) uses purity threshold $0.85$, minimum
coverage of 5 samples per rule, maximum depth 3, and sparsity
$k{=}7$, sweeping $\alpha_{\text{sc}}$ and $\alpha_{\text{conf}}$ to
study the coverage--compactness trade-off, using single
representative models. An additional evaluation on
MAGIC compares against published baselines~\cite{zarlenga2021eclaire}.

Table~\ref{tab:adult_rule_extraction} reports rule extraction
results across five BNN architectures on Adult.
Covered accuracy ($84.5$--$87.2$\%) consistently exceeds each model's
test accuracy ($84.33$--$84.63$\%), confirming extracted rules are
at least as discriminative as the underlying network on covered samples.
$\alpha_{\text{sc}}$ controls a clear coverage--compactness trade-off:
at $\alpha_{\text{sc}}{=}0.5$, compact rule sets cover $77.6$--$89.5$\% of test samples with covered accuracy
$85.9$--$87.2$\%; at $\alpha_{\text{sc}}{=}0.1$, coverage rises to
$89.9$--$95.8$\%, but with more rules and slightly lower covered accuracy
($84.5$--$86.3$\%). Total accuracy Acc$_t$ remains within
$1.2$--$2.6$\,pp of the underlying BNN, with the gap widening at lower
coverage settings where more samples fall to the CART fallback.
Sensitivity analysis of $\alpha_{\text{sc}}$, $\alpha_{\text{conf}}$,
and $k$ as interpretable trade-off knobs, and an ablation of
fallback strategies, are provided in Section~\ref{sec:ablation}.

\begin{table}[t]
  \centering
  \caption{Rule extraction on Adult across five BNN
  architectures. $\alpha_{\text{conf}}$: conflict penalty constant = 0.1. CART fallback. $\alpha_{\text{sc}}$: same-coverage penalty;
  $\overline{\ell}$: avg.\ conditions per rule;
  Cov.: coverage; Cov.\ Acc.: covered accuracy;
  Acc$_t$: total accuracy.}
  \label{tab:adult_rule_extraction}
  \vspace{2pt}
  \renewcommand{\arraystretch}{1.02}
  \setlength{\tabcolsep}{2.3pt}
  \scriptsize
  \resizebox{\columnwidth}{!}{%
  \begin{tabular}{lccccccc}
    \toprule
    Arch & BNN Acc.\ (\%) & $\alpha_{\text{sc}}$ & Rules & $\overline{\ell}$
      & Cov.\ (\%) & Cov.\ Acc.\ (\%) & Acc$_t$ (\%) \\
    \midrule
    \multirow{2}{*}{$\{14,16,2\}$}
      & \multirow{2}{*}{84.33}
      & 0.5 & 37 & 1.76 & 78.2 & \textbf{86.8} & \textbf{82.6} \\
      && 0.1 & 45 & 1.67 & \textbf{91.5} & 84.5 & 81.7 \\
    \midrule
    \multirow{2}{*}{$\{14,32,2\}$}
      & \multirow{2}{*}{84.42}
      & 0.5 & 45 & 2.04 & 84.9 & \textbf{87.2} & \textbf{83.2} \\
      && 0.1 & 53 & 2.02 & \textbf{94.0} & 85.3 & 82.1 \\
    \midrule
    \multirow{2}{*}{$\{14,128,2\}$}
      & \multirow{2}{*}{84.63}
      & 0.5 & 61 & 1.64 & 89.5 & \textbf{85.9} & \textbf{82.8} \\
      && 0.1 & 85 & 1.75 & \textbf{95.8} & 85.3 & 82.0 \\
    \midrule
    \multirow{2}{*}{$\{14,16,8,2\}$}
      & \multirow{2}{*}{84.33}
      & 0.5 & 29 & 1.69 & 84.0 & \textbf{86.9} & \textbf{83.1} \\
      && 0.1 & 39 & 1.77 & \textbf{89.9} & 86.3 & 82.3 \\
    \midrule
    \multirow{2}{*}{$\{14,32,16,2\}$}
      & \multirow{2}{*}{84.36}
      & 0.5 & 55 & 2.02 & 77.6 & \textbf{87.2} & \textbf{82.8} \\
      && 0.1 & 66 & 1.95 & \textbf{91.3} & 85.1 & 81.8 \\
    \bottomrule
  \end{tabular}
  }
\end{table}

\noindent \textbf{$\bullet$ Hardware Comparisons.}
Table~\ref{tab:hw_bern_vs_rules_dense} compares LUT-based BNNs against rule networks on hardware. The rule network consumes fixed resources ($30$ DSPs) regardless of architecture, while
LUT networks scale with width, reaching $301$ DSPs at $\{14,128,2\}$. At this largest architecture, the rule network
reduces DSP count by $10{\times}$ and latency by $1.35{\times}$, at a
cost of $1.87$\,pp in total accuracy. At smaller architectures, DSP savings persist, but
rule-based latency exceeds LUT inference as the fixed rule-matching
overhead dominates. Rule extraction thus trades a modest accuracy
cost for interpretability: predictions are expressed as conjunctions
over original input features, enabling direct inspection of decision
logic without hardware overhead scaling with model size.

\begin{table}[t]
\centering
\caption{LUT-based BNNs vs.\ rule networks HW results on Adult (AMD KV260).}
\label{tab:hw_bern_vs_rules_dense}
\scriptsize
\setlength{\tabcolsep}{2.8pt}
\begin{tabular}{llrrrrrr}
\toprule
Arch & Method & Acc (\%) & Latency & DSP & BRAM & LUT & FF \\
\midrule
\multirow{2}{*}{$\{14,16,2\}$}
  & LUT   & 84.33 & 77 & 77 & 2 & 4,329 & 5,582 \\
  & Rules & \review{82.59} & \review{183} & \review{30} & \review{5} & \review{2,997} & \review{2,290} \\ % sca=0.5 ca=0.1
\midrule
\multirow{2}{*}{$\{14,32,2\}$}
  & LUT   & 84.42 & 119 & 109 & 4 & 6,019 & 8,333 \\
  & Rules & \review{83.16} & \review{215} & \review{30} & \review{8} & \review{3,009} & \review{2,218} \\ % sca=0.5 ca=0.1
\midrule
\multirow{2}{*}{$\{14,128,2\}$}
  & LUT   & 84.63 & 376 & 301 & 28 & 15,609 & 22,865 \\
  & Rules & \review{82.76} & \review{279} & \review{30} & \review{8} & \review{3,002} & \review{2,217} \\ % sca=0.5 ca=0.1
\midrule
\multirow{2}{*}{$\{14,16,8,2\}$}
  & LUT   & 84.33 & 108 & 96 & 3 & 5,715 & 6,912 \\
  & Rules & \review{83.12} & \review{151} & \review{30} & \review{5} & \review{3,011} & \review{2,300} \\ % sca=0.5 ca=0.1
\midrule
\multirow{2}{*}{$\{14,32,16,2\}$}
  & LUT   & 84.36 & 172 & 144 & 6 & 8,352 & 10,990 \\
  & Rules & \review{82.78} & \review{255} & \review{30} & \review{8} & \review{3,027} & \review{2,218} \\ % sca=0.5 ca=0.1
\bottomrule
\end{tabular}
\end{table}

% ============================================================
\noindent \textbf{$\bullet$ Comparison to Prior Methods.}
\label{sec:results_magic}
Table~\ref{tab:magic} compares symbolic rule extraction on MAGIC
against published baselines~\cite{zarlenga2021eclaire} under
5-fold cross-validation, using ECLAIRE's reference ReLU MLP $\{10,64,32,16,2\}$ as the teacher, and evaluating two settings: (i) rule
extraction from a BNN trained from scratch on the same architecture
(no KD), and (ii) rule extraction from a compressed BNN student $\{10,64,32,2\}$ distilled from that teacher ($T{=}2$, $\alpha{=}0.5$).

The extraction pipeline uses a purity threshold of $0.85$, sparsity $k{=}3$, and maximum rule depth of 2. 
The KD configuration matches ECLAIRE's accuracy ($84.4$\% vs.\
$84.6$\%) with $9{\times}$ fewer rules ($44{\pm}4$ vs.\ $396{\pm}75$)
and half the average rule complexity ($1.86$ vs.\ $3.82$ conditions), while achieving higher fidelity to the underlying network ($91.8$\% vs.\ $89.4$\%). Without KD, rule sets remain compact ($24$--$46$ rules) at comparable accuracy ($82.9$--$84.2$\%), indicating rule compactness is primarily attributable to Bernstein activation geometry rather than distillation, with KD contributing a small gain in accuracy or fidelity.

\begin{table}[t]
  \centering
  \caption{Rule extraction on MAGIC Gamma Telescope.
  Baselines from Zarlenga et al.~\cite{zarlenga2021eclaire} (mean\,$\pm$\,std, 5 folds); Decompositional: rule extraction from intermediate
  layers; $\overline{\ell}$: avg.\ conditions per rule; Fid: fidelity
  to baseline network.}
  \label{tab:magic}
  \vspace{2pt}
  \renewcommand{\arraystretch}{1.02}
  \setlength{\tabcolsep}{3.5pt}
  \scriptsize
  \begin{tabular}{llcccc}
    \toprule
    Type & Method & Rules & $\overline{\ell}$ & Acc (\%) & Fid (\%) \\
    \midrule
    Decompositional & DeepRED~\cite{zilke2016deepred}
      & $5143\pm9799$ & 5.43 & 78.7 & 89.3 \\
    & REM-D~\cite{shams2021rem}
      & $3617\pm6748$ & 5.41 & 78.6 & 89.4 \\
    & ECLAIRE~\cite{zarlenga2021eclaire}
      & $396\pm75$    & 3.82 & \textbf{84.6} & 89.4 \\
    \midrule
    \sysname & no-KD ($\alpha_{\text{sc}}{=}0.1$)
      & \review{$46\pm4$} & \review{1.84} & \review{84.2} & \review{91.8} \\
    & no-KD ($\alpha_{\text{sc}}{=}0.5$)
      & \review{$\mathbf{24\pm3}$} & \review{$\mathbf{1.81}$} & \review{82.9} & \review{94.3} \\
    & KD ($\alpha_{\text{sc}}{=}0.1$)
      & \review{$44\pm4$} & \review{$1.86$} & \review{$\mathbf{84.4}$} & \review{$91.8$} \\
    & KD ($\alpha_{\text{sc}}{=}0.5$)
      & \review{$\mathbf{24\pm4}$} & \review{$1.87$} & \review{$82.8$} & \review{$\mathbf{94.6}$} \\
    \bottomrule
  \end{tabular}
\end{table}

\subsection{\sysname End-to-End Results}
Table~\ref{tab:bern2edge_hw} presents end-to-end post-synthesis
results for \sysname against the quantized teacher baseline across all
three datasets. \sysname (LUT) reduces latency by $91.9$--$99.8$\% relative
to the W8A8 teacher while matching or staying within $0.5$\,pp of teacher
accuracy. DSP savings reach $27.4$\% on HIGGS-Small and $74.7$\% on Adult.
On Covertype, the apparent BRAM increase ($128 \rightarrow 149$) reflects a
memory-architecture shift: the teacher relies on 52 URAM tiles for weight
storage, whereas \sysname substitutes standard BRAM and eliminates $94.2$\%
of URAM usage ($52 \rightarrow 3$), reducing DSPs by $16.2$\%.

On Adult, the rule deployment path cuts DSP usage to 30
($\downarrow 89.0$\%) relative to the teacher, at a cost of $1.5$\,pp in
total accuracy compared to the LUT path. Beyond resource savings, rule
deployment constrains predictions to explicit conjunctions over the original
input features, providing formal guarantees on the input subspace covered by
each decision; a property unavailable in either the teacher model or the
LUT-based BNN.

\begin{table}[t]
\centering
\caption{Post-synthesis accuracy and end-to-end hardware results for \sysname across three datasets. Percentages indicate change relative to the W8A8 teacher baseline. Rule deployment path reported for Adult only.}
\label{tab:bern2edge_hw}
\renewcommand{\arraystretch}{1.0}
\setlength{\tabcolsep}{0.7pt}
\scriptsize
\begin{tabular}{l c r r r r}
\toprule
Method & Acc (\%) & Latency (cycles) & DSPs & BRAMs & URAMs \\ 
\midrule
\multicolumn{6}{l}{\textit{HIGGS-Small}} \\
\midrule
Teacher (W8A8)               & 72.3 & 6,981 & 186 & 50& -- \\
Bern2Edge (LUT)        & 72.3 & 125 ($\downarrow$ 98.2\%)  & 135 ($\downarrow$ 27.4\%)  & 3 ($\downarrow$ 94.0\%)& -- \\
\midrule
\multicolumn{6}{l}{\textit{Covertype}} \\
\midrule
Teacher (W8A8)                & 96.9 & 91,634 & 389 & 128 & 52\\ 
Bern2Edge (LUT) & 96.4 & 7,433 ($\downarrow$ 91.9\%)
& 326 ($\downarrow$ 16.2\%)
& 149 ($\uparrow$ 16.4\%)
& 3 ($\downarrow$ 94.2\%)\\
\midrule
\multicolumn{6}{l}{\textit{Adult}} \\
\midrule
Teacher (W8A8)               & 84.6 & 40,305 & 273 & 42 & 20\\
Bern2Edge (LUT)        & 84.6 & 70 ($\downarrow$ 99.8\%)  & 69 ($\downarrow$ 74.7\%)  & 2 ($\downarrow$ 95.2\%)& -- \\
Bern2Edge (Rules)      & \review{83.12} & \review{151 ($\downarrow$ 99.6\%)}  & \review{30 ($\downarrow$ 89.0\%)}  & \review{5 ($\downarrow$ 88.1\%)} & --\\
\bottomrule
\end{tabular}
\end{table}

\subsection{Deployment and Optimization on Low-Power Edge}
We evaluate both \sysname deployment paths on severely
resource-constrained hardware. The Spartan-7 XC7S15~\cite{xilinx_spartan7_ds189},
a low-power FPGA, provides only 8\,k LUTs, 20 DSPs, and 20 BRAM18K
at ${\sim}16$\,mW static power, $14.6{\times}$, $62{\times}$, and
$14.4{\times}$ fewer LUTs, DSPs, and BRAMs than the KV260, respectively.
We deploy LUT-based BNNs on Adult using the $\{14,h,2\}$ architecture over $h\!\in\!\{4,8,16,32,64,128\}$ hidden layer sizes, and symbolic rule networks, across two
rule-count configurations.
Two device-level optimizations bring each path within budget
without altering any model parameters. For the rule system, \emph{distributed-RAM rule storage} binds the
3.5\,KB INT8 rule ROM to LUTRAM rather than block RAM, bringing it
within the BRAM budget. For the BNN,
\emph{fixed-point quantization} reduces the linear-layer word from \texttt{fix<32,16>} to \texttt{fix<18,8>}, mapping each
linear-layer multiply to a single DSP rather than three, bringing every configuration within the DSP budget at a cost of ${\le}0.1$\,pp accuracy. The bitwidth split reflects two separate
constraints: 8 integer bits are required to represent Adult's
categorical codes (up to 40), while the remaining 10 fractional
bits bound quantization rounding error and recover accuracy to
within $0.1$\,pp of fp32 at no added DSP cost.
Table~\ref{tab:lp_deploy} reports post-synthesis results across all
configurations. Every design fits comfortably within device limits,
and DSP count stays nearly flat across all widths due to the time-multiplexed II$=$1 linear-layer datapath, so wider networks
pay primarily in latency and activation BRAM rather than
combinational logic. Rule-based classifiers minimize DSP
usage, while the smallest Bernstein configurations (e.g., $h{=}4$)
achieve lower LUT, BRAM, latency, and power, making symbolic
deployment resource-optimal specifically when DSP is the binding
constraint.

\begin{table}[t]
\centering
\caption{Deployment on XC7S15. R50 and R29 denote the
50-rule and 29-rule symbolic classifiers, respectively.
Acc is test accuracy; Acc$_{\mathrm{fp}}$ the fp32 reference for BNNs.}
\label{tab:lp_deploy}
\scriptsize
\setlength{\tabcolsep}{1.9pt}
\resizebox{\columnwidth}{!}{%
\begin{tabular}{llcccccccc}
\toprule
Model & Config. & LUT & FF & DSP & BRAM & Lat.\ & Pwr & Acc & Acc$_{\mathrm{fp}}$ \\
         &        &     &    &     & 18K  & (cyc) & (mW) & (\%) & (\%) \\
\midrule
\multirow{2}{*}{Rules}
 & R50 & 1{,}086 & 844 & 16 & 6 & 340 & 62 & 83.02 & --- \\
 & R29 & \textbf{710} & 765 & 16 & 4 & 156 & \textbf{59} & 83.01 & --- \\
\midrule
\multirow{6}{*}{\makecell[l]{Bern.\\$\{14,h,2\}$}}
 & $h{=}4$   & 553 & 883 & 18 & 1  & 61  & 51 & 84.18 & 84.25 \\
 & $h{=}8$   & 635 & 804 & 19 & 1  & 82  & 63 & 84.57 & 84.64 \\
 & $h{=}16$  & 662 & 777 & 19 & 1  & 115 & 63 & 84.88 & 84.86 \\
 & $h{=}32$  & 667 & 799 & 19 & 2  & 179 & 58 & 84.92 & 84.92 \\
 & $h{=}64$  & 698 & 826 & 19 & 5  & 308 & 60 & 84.82 & 84.92 \\
 & $h{=}128$ & 858 & 870 & 19 & 13 & 565 & 66 & 84.95 & 85.03 \\
\bottomrule
\end{tabular}
}
\end{table}

\subsection{Hyperparameter and Fallback Ablation}
\label{sec:ablation}

Bern2Edge exposes three parameters governing rule extraction: the
sparsity threshold~$k$, the redundancy penalty~$\alpha_{\text{sc}}$,
and the conflict penalty~$\alpha_{\text{conf}}$. Each controls a
specific hardware--accuracy trade-off with monotone, predictable
behavior. The rule purity threshold is set close to the underlying
network's accuracy, ensuring extracted rules are at least as
discriminative as the model they summarize.

\begin{figure}[t]
  \centering
  \begin{tikzpicture}
\begin{axis}[
    width=0.9\columnwidth,
    height=4.2cm,
    xlabel={sparsity $k$},
    ylabel={BRAM},
    ylabel style={color=blue!70!black},
    yticklabel style={color=blue!70!black},
    axis y line*=left,
    axis x line*=bottom,
    xtick={1,2,3,4,5,6,7,8,9,10,11,12,13},
    xmin=0.5, xmax=13.5,
    ymin=0, ymax=16,
    grid=both,
    grid style={line width=0.4pt, draw=gray!30},
    tick label style={font=\small},
    label style={font=\small},
    name=mainax,
]
\addplot[
    color=blue!70!black,
    solid,
    mark=*,
    mark size=2pt,
    line width=0.9pt,
] coordinates {
    (1,  2)
    (2,  3)
    (3,  4)
    (4,  5)
    (5,  6)
    (6,  7)
    (7,  8)
    (8,  9)
    (9,  10)
    (10, 11)
    (11, 12)
    (12, 13)
    (13, 14)
};
\end{axis}

\begin{axis}[
    width=0.9\columnwidth,
    height=4.2cm,
    ylabel={Total Acc. (\%)},
    ylabel style={color=red!70!black},
    yticklabel style={color=red!70!black},
    axis y line*=right,
    axis x line=none,
    xtick={1,2,3,4,5,6,7,8,9,10,11,12,13},
    xmin=0.5, xmax=13.5,
    ymin=76, ymax=83,
    tick label style={font=\small},
    label style={font=\small},
    name=rightax,
]
\addplot[
    color=red!70!black,
    solid,
    mark=square*,
    mark size=2pt,
    line width=0.9pt,
] coordinates {
    (1,  77.67)
    (2,  79.43)
    (3,  79.49)
    (4,  79.02)
    (5,  80.65)
    (6,  81.08)
    (7,  81.96)
    (8,  81.24)
    (9,  81.77)
    (10, 81.27)
    (11, 81.42)
    (12, 80.56)
    (13, 81.99)
};
\end{axis}
\end{tikzpicture}
  \caption{
    BRAM and total accuracy vs.\ sparsity
    threshold~$k$ on Adult ($h=32$).
  }
  \label{fig:k_sweep}
\end{figure}
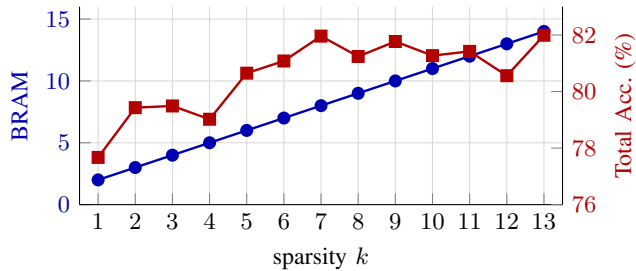

\noindent \textbf{$\bullet$ Sparsity threshold $k$}
$k$ controls per-rule memory footprint only, with no effect on rule
count. Fig.~\ref{fig:k_sweep} shows BRAM grows linearly with $k$ while
total accuracy plateaus beyond $k{=}7$ with increasing variance for
$k{>}7$. We select $k{=}7$, achieving ${\sim}55$\% of the dense
memory footprint ($k{=}14$) at no accuracy cost. This plateau reflects
that most rule weights are small and can be zeroed without loss; sparse
rules match dense accuracy because the dropped conditions carry little
discriminative weight.

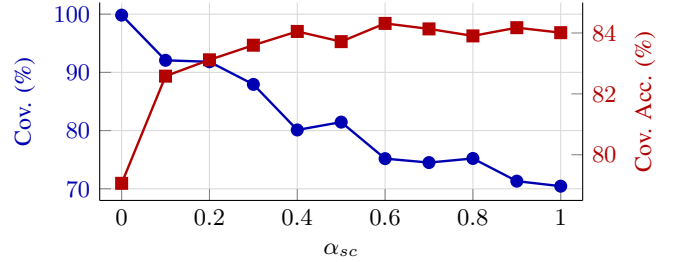
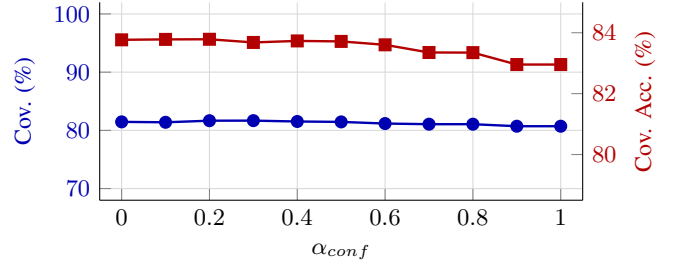
\begin{figure}[t]
  \centering 
  \subfloat[Sweep $\alpha_{\text{sc}}$ ($\alpha_{\text{conf}}=0.5$)]{\begin{tikzpicture}

\begin{axis}[
    name=alphaScAxis,
    width=0.9\columnwidth,
    height=4.2cm,
    title style={font=\small, yshift=-2pt},
    xlabel={$\alpha_{sc}$},
    ylabel={Cov. (\%)},
    ylabel style={color=blue!70!black},
    yticklabel style={color=blue!70!black},
    axis y line*=left,
    axis x line*=bottom,
    xtick={0,0.2,0.4,0.6,0.8,1.0},
    xmin=-0.05, xmax=1.05,
    ymin=68, ymax=102,
    grid=both,
    grid style={line width=0.4pt, draw=gray!30},
    tick label style={font=\small},
    label style={font=\small},
]
\addplot[
    color=blue!70!black,
    solid,
    mark=*,
    mark size=2pt,
    line width=0.9pt,
] coordinates {
    (0.0, 99.832)
    (0.1, 92.072)
    (0.2, 91.820)
    (0.3, 87.936)
    (0.4, 80.108)
    (0.5, 81.454)
    (0.6, 75.180)
    (0.7, 74.506)
    (0.8, 75.210)
    (0.9, 71.312)
    (1.0, 70.450)
};
\end{axis}

\begin{axis}[
    at={(alphaScAxis.south east)},
    anchor=south east,
    width=0.9\columnwidth,
    height=4.2cm,
    ylabel={Cov. Acc. (\%)},
    ylabel style={color=red!70!black},
    yticklabel style={color=red!70!black},
    axis y line*=right,
    axis x line=none,
    xtick=\empty,
    xmin=-0.05, xmax=1.05,
    ymin=78.5, ymax=85,
    tick label style={font=\small},
    label style={font=\small},
]
\addplot[
    color=red!70!black,
    solid,
    mark=square*,
    mark size=2pt,
    line width=0.9pt,
] coordinates {
    (0.0, 79.062)
    (0.1, 82.582)
    (0.2, 83.118)
    (0.3, 83.598)
    (0.4, 84.050)
    (0.5, 83.716)
    (0.6, 84.314)
    (0.7, 84.134)
    (0.8, 83.904)
    (0.9, 84.174)
    (1.0, 84.010)
};
\end{axis}

\end{tikzpicture}\label{fig:sc_sweep}}
  \hfill
  \subfloat[Sweep $\alpha_{\text{conf}}$ ($\alpha_{\text{sc}}=0.5$)]{\begin{tikzpicture}

\begin{axis}[
    name=alphaConfAxis,
    width=0.9\columnwidth,
    height=4.2cm,
    title style={font=\small, yshift=-2pt},
    xlabel={$\alpha_{conf}$},
    ylabel={Cov. (\%)},
    ylabel style={color=blue!70!black},
    yticklabel style={color=blue!70!black},
    axis y line*=left,
    axis x line*=bottom,
    xtick={0,0.2,0.4,0.6,0.8,1.0},
    xmin=-0.05, xmax=1.05,
    ymin=68, ymax=102,
    grid=both,
    grid style={line width=0.4pt, draw=gray!30},
    tick label style={font=\small},
    label style={font=\small},
]
\addplot[
    color=blue!70!black,
    solid,
    mark=*,
    mark size=2pt,
    line width=0.9pt,
] coordinates {
    (0.0, 81.460)
    (0.1, 81.386)
    (0.2, 81.662)
    (0.3, 81.672)
    (0.4, 81.524)
    (0.5, 81.454)
    (0.6, 81.172)
    (0.7, 81.060)
    (0.8, 81.060)
    (0.9, 80.712)
    (1.0, 80.712)
};
\end{axis}

\begin{axis}[
    at={(alphaConfAxis.south east)},
    anchor=south east,
    width=0.9\columnwidth,
    height=4.2cm,
    ylabel={Cov. Acc. (\%)},
    ylabel style={color=red!70!black},
    yticklabel style={color=red!70!black},
    axis y line*=right,
    axis x line=none,
    xtick=\empty,
    xmin=-0.05, xmax=1.05,
    ymin=78.5, ymax=85,
    tick label style={font=\small},
    label style={font=\small},
]
\addplot[
    color=red!70!black,
    solid,
    mark=square*,
    mark size=2pt,
    line width=0.9pt,
] coordinates {
    (0.0, 83.768)
    (0.1, 83.782)
    (0.2, 83.786)
    (0.3, 83.682)
    (0.4, 83.732)
    (0.5, 83.716)
    (0.6, 83.604)
    (0.7, 83.352)
    (0.8, 83.348)
    (0.9, 82.956)
    (1.0, 82.956)
};
\end{axis}

\end{tikzpicture}\label{fig:conf_sweep}}
  
  \caption{
    Coverage and covered accuracy as functions of the two rule-extraction penalty parameters on Adult, averaged across five network architectures (\(h\in\{16,32,128,16{\times}8,32{\times}16\}\)).
  }
  \label{fig:param_sweep}
\end{figure}

\noindent \textbf{$\bullet$ Penalty parameters}
Table~\ref{tab:adult_rule_sweep} details the full metric picture
(conflict count, rule count, and average conditions per rule)
across joint $(\alpha_{\text{sc}}, \alpha_{\text{conf}})$ settings.
Fig.~\ref{fig:param_sweep} isolates each parameter's effect on coverage
and covered accuracy. $\alpha_{\text{sc}}$ is the primary lever:
increasing it from $0$ to $1$ trades coverage (${\sim}100$\%
${\to}{\sim}70$\%) for covered accuracy (${\sim}79$\%${\to}{\sim}84$\%),
with an elbow near $\alpha_{\text{sc}}{\approx}0.2$. This is a deployment
choice, not a sensitivity: broad-coverage applications favor small
$\alpha_{\text{sc}}$; high-confidence rule firing favors large
$\alpha_{\text{sc}}$. Within each $\alpha_{\text{sc}}$ regime,
$\alpha_{\text{conf}}$ reduces raw conflict count
(Table~\ref{tab:adult_rule_sweep}) but has negligible effect on coverage
or covered accuracy ($\lesssim$0.6~pp variation across the full range), since
conflicts are resolved at inference time by firing the highest-purity
matching rule. $\alpha_{\text{conf}}$ therefore requires no careful tuning.

\begin{table}[t]
\centering 
\caption{Joint penalty sweep on Adult ($h=32$, BNN acc.\ $84.42$\%).
Conf.: conflicting samples; Cov.: coverage;
Cov.\ Acc.: covered accuracy; Acc$_t$: total accuracy;
$\overline{\ell}$: avg.\ conditions per rule.}
\label{tab:adult_rule_sweep}
\renewcommand{\arraystretch}{1.02}
\scriptsize
\setlength{\tabcolsep}{2.5pt}
\begin{tabular}{cc cccccc}
\toprule
$\alpha_{\text{conf}}$ & $\alpha_{\text{sc}}$
  & Conf. & Cov.\ (\%) & Cov.\ Acc.\ (\%) & Acc$_t$ (\%) & Rules & $\overline{\ell}$ \\
\midrule
0.1 & 0.1 & 308 & \textbf{94.01} & \textbf{85.26} & 82.10 & 53 & \textbf{2.02} \\
1.0 & 0.1 & \textbf{156} & 92.60 & 84.67 & \textbf{82.39} & \textbf{50} & 2.06 \\
\midrule
0.1 & 0.3 & 136 & \textbf{87.15} & 86.72 & \textbf{83.16} & 53 & 2.06 \\
1.0 & 0.3 & \textbf{111} & 86.99 & \textbf{86.77} & \textbf{83.16} & \textbf{52} & \textbf{2.04} \\
\midrule
0.1 & 0.5 & 121 & \textbf{84.86} & \textbf{87.18} & 83.16 & \textbf{45} & \textbf{2.04} \\
1.0 & 0.5 & \textbf{106} & 84.42 & \textbf{87.18} & \textbf{83.39} & 46 & 2.09 \\
\midrule
0.1 & 1.0 & 78 & \textbf{82.40} & 87.23 & 83.43 & 37 & 1.95 \\
1.0 & 1.0 & \textbf{70} & 80.69 & \textbf{87.61} & \textbf{83.46} & \textbf{36} & \textbf{1.94} \\
\bottomrule
\end{tabular}
\end{table}

\noindent \textbf{$\bullet$ Fallback Ablation}
Table~\ref{tab:fallback_ablation} compares four fallback strategies on
the Adult dataset across accuracy, fidelity to the underlying network
on uncovered samples, and hardware cost. The linear regression
fallback operates in first-layer pre-activation space
$\mathbf{z} = W_0 \mathbf{x} + \boldsymbol{\beta}_0$, folded into an equivalent input-space classifier $\mathbf{w}_{\text{eff}}^\top \mathbf{x} + \beta_{\text{eff}}$.
The small BNN fallback uses a one-hidden-layer network of width~4
trained on uncovered samples only; the full BNN is the original
network, so its fidelity is 100\% by definition.
We select CART (depth $\leq 4$) as the fallback: it achieves the best
overall accuracy (83.02\%) and the highest accuracy on uncovered
samples (72.1\%) at the lowest hardware cost (714 LUT, 471 FF, 0 DSP).
Notably, even the full BNN matches neither metric, suggesting uncovered
samples are inherently difficult to classify regardless of model
capacity. The full BNN also requires $2.6{\times}$ more LUTs and
$4.9{\times}$ more flip-flops than CART. Linear regression offers
the lowest latency (13 cycles) but yields the worst accuracy on
uncovered samples (67.4\%), reflecting the limitations of a linear
decision boundary in this region.

\begin{table}[t] 
\caption{Fallback strategy ablation on the Adult dataset ($h=32$, $\alpha_{\text{sc}}=0.5$, $\alpha_{\text{conf}}=0.1$).
Resource columns report the fallback component only.
Acc$_{\text{unc}}$ and Fid$_{\text{unc}}$ are on uncovered samples.
All accuracy and fidelity values in \%.}
\label{tab:fallback_ablation}
\centering
\scriptsize
\setlength{\tabcolsep}{2.8pt}
\renewcommand{\arraystretch}{1.1}
\resizebox{\columnwidth}{!}{%
\begin{tabular}{lcccccccc}
\toprule
\textbf{Fallback} & \textbf{Acc.} & \textbf{Acc$_{\text{unc}}$} &
\textbf{Fid$_{\text{unc}}$} & \textbf{LUT} & \textbf{FF} &
\textbf{DSP} & \textbf{Lat.} \\
\midrule
Linear Regression & 81.67 & 67.4 & 84.1  & 1749 & 520  & 10 & \textbf{13} \\
Full BNN            & 82.91 & 71.9 & \textbf{100.0} & 1887 & 2318 & 40 & 68 \\
Small BNN           & 82.23 & 71.3 & 90.5  & 1129 & 2178 & 38 & 46 \\
CART                & \textbf{83.02} & \textbf{72.1} & 87.7 & \textbf{714} & \textbf{471} & \textbf{0} & 26 \\
\bottomrule
\end{tabular}%
}
\end{table}

\subsection{Symbolic Rule Certification and Robustness}

\noindent \textbf{$\bullet$ Rule Certification on Input Noise.}
We evaluate the symbolic rule system on MAGIC, using a $\{10,64,32,2\}$
architecture for the ReLU network and BNN, and the rules extracted from
that BNN. Inputs are perturbed by bounded per-feature noise with budget
$\epsilon_i = c \cdot s_i$, where $s_i$ is the training-set standard
deviation of feature~$i$. The range $c \leq 0.05$ corresponds to
sub-$5$\% relative uncertainty, consistent with calibrated process
measurement devices~\cite{iec61298}; $c = 0.10$ serves as a stress test
beyond normal operating conditions.

For each rule slab $z^{\text{lo}} \leq \mathbf{w}^\top \mathbf{x} < z^{\text{hi}}$, the raw input box $[x_i - \epsilon_i, x_i + \epsilon_i]$ is mapped through the monotone quantile normalization, and the exact worst-case interval of $\mathbf{w}^\top \mathbf{x}$ is propagated over the box. A point is \emph{rule
certified} if no different-label rule of equal or higher purity is
reachable within the box, i.e.\ no admissible perturbation can change the
predicted label. This certificate is conservative and therefore never
falsely certifies stability.

We compare rule certificates against network-level certificates for the underlying
BNN and equivalent ReLU network. The BNN uses
Bern-IBP with de~Casteljau subdivision~\cite{khedr2024deepbern}; the ReLU
network uses standard IBP via auto\_LiRPA~\cite{xu2020automatic}, following
the comparison methodology of Khedr et al.~\cite{khedr2024deepbern}.
Empirical robustness is estimated with $20$ i.i.d.\ uniform perturbations
per point from the same noise box (${\approx}50{,}000$ perturbed inputs per
noise level). Since uniform sampling rarely reaches box corners, empirical
robustness is an upper bound; certified columns are sound lower bounds.

Table~\ref{tab:robustness} reports results on the covered-correct test subset
(2{,}485 of MAGIC test points). Empirical rule accuracy drops by only
3.0\,pp at $c = 0.10$, and rule--BNN fidelity remains above 95\% for all
realistic noise levels ($c \leq 0.05$). The certified results reveal a larger
separation: at $c = 0.05$, rule certification reaches 42.7\%, versus
16.0\% for the BNN and 12.4\% for the matched ReLU network. This advantage stems from certifying compact linear slab rules
directly; the resulting intervals are tighter than those produced by network-level IBP at every noise level tested. Among the neural models, the BNN achieves higher certified
stability than the matched ReLU network at every noise level,
consistent with the tighter bounds produced by
Bern-IBP~\cite{khedr2024deepbern}.

\begin{table}[t]
\centering
\caption{\review{Robustness certification on MAGIC. Empirical columns show
rule accuracy and rule--BNN fidelity under sampled noise; certified columns
show provably stable fractions for rules and networks.}}
\label{tab:robustness}
\scriptsize
\renewcommand{\arraystretch}{0.8}
\resizebox{\columnwidth}{!}{%
\begin{tabular}{cccccc}
\toprule
& \multicolumn{2}{c}{Empirical (\%)} & \multicolumn{3}{c}{Certified (\%)} \\
\cmidrule(lr){2-3}\cmidrule(lr){4-6}
$c$ & Rule Acc. & Fid.
& Rules
& BNN & ReLU \\
\midrule
0.00 & 100.0 & 98.2 & 100.0 & 100.0 & 100.0 \\
0.01 &  99.6 & 97.8 &  \textbf{83.4}&  82.1 &  80.1 \\
0.03 &  99.0 & 97.2 & \textbf{59.4} &  41.3 &  27.1 \\
0.05 &  98.4 & 96.4 &  \textbf{42.7}&  16.0 &  12.4 \\
0.10 &  97.0 & 94.7 &  \textbf{22.7} &   6.6 &   4.8 \\
\bottomrule
\end{tabular}%
}
\end{table}

%==============distribution shift==================
% ── PROSE PARAGRAPH ──────────────────────────────────────────────

\noindent \textbf{$\bullet$ Rule Robustness to Data Distribution Shift.}
A key concern for streaming edge deployment is whether rules remain reliable under distribution shift.
Table~\ref{tab:dist_shift} evaluates this on ACS~Income~\cite{ding2021retiring},
training on CA-2018 and testing under temporal and geographic shifts. The
ReLU teacher uses a $\{10,512,256,128,2\}$ architecture; the ReLU and BNN
students both use $\{10,32,2\}$, with rules extracted from the BNN student.

Under temporal shift, rule coverage is essentially stable ($+0.2$~pp) and
covered accuracy drops by only 1.6~pp, closely tracking the underlying BNN's own 1.1~pp accuracy loss. Geographic shift is harder:
coverage falls 4.3~pp but remains above 85\%, and covered accuracy drops
3.9~pp, comparable to the 6.4--7.0~pp accuracy degradation of the neural
baselines. In both conditions, symbolic extraction does not amplify the
underlying model's degradation. Crucially, uncovered samples are explicitly
flagged at deployment and routed to the fallback model, so distributional
failures surface as measurable coverage drops rather than silent mispredictions.

\begin{table*}[!ht]
\caption{Distribution-shift robustness on ACS~Income, training on CA-2018.
ID = held-out CA-2018 test set; GEO-AVG and TEMP-AVG are
means over shifted conditions; $\Delta$ = AVG$-$ID (pp).
Means$\pm$std over 5 seeds.}
\label{tab:dist_shift}
\centering
\scriptsize
\renewcommand{\arraystretch}{1.1}
\resizebox{\textwidth}{!}{%
\begin{tabular}{ll c ccccc ccccc}
\toprule
& & &
  \multicolumn{5}{c}{\textbf{Geographic shift}} &
  \multicolumn{5}{c}{\textbf{Temporal shift (CA)}} \\
\cmidrule(lr){4-8}\cmidrule(lr){9-13}
\textbf{System} & \textbf{Metric} &
  \textbf{ID (CA-18)} &
  \textbf{GEO-AVG} & $\Delta$ & MS & WY & WV &
  \textbf{TEMP-AVG} & $\Delta$ & 2019 & 2021 & 2022 \\
\midrule
ReLU Teacher &
  Acc.\ (\%) &
  81.28\std{0.09} & 74.32\std{0.64} & $-$7.0 &
  73.73 & 74.61 & 74.61 &
  80.00\std{0.09} & $-$1.3 &
  80.74 & 79.82 & 79.44 \\[2pt]
ReLU &
  Acc.\ (\%) &
  80.55\std{0.15} & 74.10\std{0.76} & $-$6.5 &
  73.42 & 74.45 & 74.43 &
  79.37\std{0.19} & $-$1.2 &
  80.03 & 79.23 & 78.84 \\[2pt]
BNN &
  Acc.\ (\%) &
  80.89\std{0.29} & 74.45\std{0.69} & $-$6.4 &
  73.79 & 74.77 & 74.79 &
  79.81\std{0.36} & \textbf{$-$1.1} &
  80.49 & 79.67 & 79.26 \\
\midrule
\multirow{3}{*}{\shortstack[l]{Rules}} &
  Coverage (\%) &
  90.09\std{0.86} & 85.82\std{1.38} & $-$4.3 &
  86.10 & 85.07 & 86.29 &
  90.24\std{0.78} & $+$0.2 &
  90.01 & 90.35 & 90.35 \\
&
  Covered acc.\ (\%) &
  80.29\std{0.20} & 76.41\std{1.07} & \textbf{$-$3.9} &
  77.31 & 74.82 & 77.09 &
  78.67\std{0.42} & $-$1.6 &
  79.38 & 78.62 & 78.00 \\
&
  Total acc.\ (\%) &
  78.58\std{0.12} & 73.15\std{1.35} & $-$5.4 &
  73.41 & 72.44 & 73.61 &
  77.24\std{0.29} & $-$1.3 &
  77.78 & 77.24 & 76.70 \\
\bottomrule
\end{tabular}}
\end{table*}

\subsection{Extending to Transformer FFN Layers}
We extend Bern2Edge to the FFN sublayers of TinyBERT4~\cite{jiao2020tinybert}, demonstrating BNN deployment in transformer 
architectures. The original TinyBERT4 FFN hidden size ($h{=}1200$) is compressed to $h \in \{312, 600\}$ across all FFN layers, with all other components unchanged. Training follows three stages: (1) function-matching from the teacher
GeLU FFN to a degree-15 BNN at the target width, where higher degree
imposes no additional hardware cost under LUT-based synthesis
(Fig.~\ref{fig:hw_resources}),
(2) substitution of the compressed BNN into the original TinyBERT4
transformer, and (3) 10 epochs of KD fine-tuning from the original
TinyBERT4 using the training scheme in Section~\ref{sec:training}.

We synthesize and implement all TinyBERT4 encoder variants on KV260. The $h{=}600$ Bernstein FFN matches or exceeds TinyBERT4 accuracy ($90.48$\% vs.\ $90.37$\%) despite halving the FFN hidden width. The Bernstein variants' reduced computational depth translates directly to latency, with end-to-end cycle count dropping by up to $61.0\%$ relative to the teacher and the FFN sublayer alone reaching $72.2\%$ (Table~\ref{tab:transformer_results}). It also reduces DSP, FF, and LUT usage relative to both the teacher and the GeLU baseline at matched width, whereas GeLU's direct polynomial computation remains at the teacher's DSP count regardless of compression. The higher BRAM usage of Bernstein FFNs relative to GeLU is attributable to per-neuron LUT storage, though both remain below TinyBERT4's original footprint. Despite their smaller size, both compressed GeLU variants increase FF usage over the teacher, as their buffers are synthesized as distributed RAM, implemented directly in FFs and LUTs rather than dedicated memory blocks.

\begin{table}[!t]
\centering
\caption{SST-2 accuracy, latency, and FPGA resource utilization for full 4-layer FFN substitution. \emph{FFN}: FFN sublayers only; \emph{Full}: complete encoder. Resource reductions are relative to TinyBERT4.}
\label{tab:transformer_results}
\scriptsize
\setlength{\tabcolsep}{2.8pt}
\renewcommand{\arraystretch}{1.4}
\resizebox{\columnwidth}{!}{%
\begin{tabular}{llccccc}
\toprule
& & \multirow{2}{*}{\textbf{TinyBERT4}}
& \multicolumn{2}{c}{\textbf{$h{=}600$}}
& \multicolumn{2}{c}{\textbf{$h{=}312$}} \\
\cmidrule(lr){4-5}\cmidrule(lr){6-7}
\textbf{Scope} & \textbf{Metric}
& & \textbf{GeLU} & \textbf{Bern} & \textbf{GeLU} & \textbf{Bern} \\
\midrule

& SST-2 Acc. (\%)
& 90.37
& 90.02
& \textbf{90.48}
& 89.11
& 90.02 \\
\cmidrule(lr){2-7}

\multirow[t]{5}{*}{FFN}
& \makecell[t]{Lat. (cycles)}
& \makecell[t]{6,413,520\\[-0.2ex]~}
& \makecell[t]{3,282,640\\($\downarrow$48.8\%)}
& \makecell[t]{3,278,288\\($\downarrow$48.9\%)}
& \makecell[t]{1,789,648\\($\downarrow$72.1\%)}
& \makecell[t]{\textbf{1,785,296}\\\textbf{($\downarrow$72.2\%)}} \\

& \makecell[t]{DSP}
& \makecell[t]{121\\[-0.2ex]~}
& \makecell[t]{121\\(0.0\%)}
& \makecell[t]{\textbf{80}\\\textbf{($\downarrow$33.9\%)}}
& \makecell[t]{121\\(0.0\%)}
& \makecell[t]{\textbf{80}\\\textbf{($\downarrow$33.9\%)}} \\

& \makecell[t]{BRAM}
& \makecell[t]{155\\[-0.2ex]~}
& \makecell[t]{82\\($\downarrow$47.1\%)}
& \makecell[t]{128\\($\downarrow$17.4\%)}
& \makecell[t]{\textbf{80}\\\textbf{($\downarrow$48.4\%)}}
& \makecell[t]{108\\($\downarrow$30.3\%)} \\

& \makecell[t]{FF}
& \makecell[t]{17,746\\[-0.2ex]~}
& \makecell[t]{19,580\\($\uparrow$10.3\%)}
& \makecell[t]{15,654\\($\downarrow$11.8\%)}
& \makecell[t]{19,528\\($\uparrow$10.0\%)}
& \makecell[t]{\textbf{15,568}\\\textbf{($\downarrow$12.3\%)}} \\

& \makecell[t]{LUT}
& \makecell[t]{34,744\\[-0.2ex]~}
& \makecell[t]{35,165\\($\uparrow$1.2\%)}
& \makecell[t]{30,070\\($\downarrow$13.5\%)}
& \makecell[t]{34,924\\($\uparrow$0.5\%)}
& \makecell[t]{\textbf{29,789}\\\textbf{($\downarrow$14.3\%)}} \\

\midrule

\multirow[t]{5}{*}{Full}
& \makecell[t]{Lat. (cycles)}
& \makecell[t]{7,591,584\\[-0.2ex]~}
& \makecell[t]{4,460,704\\($\downarrow$41.2\%)}
& \makecell[t]{4,456,352\\($\downarrow$41.3\%)}
& \makecell[t]{2,967,712\\($\downarrow$60.9\%)}
& \makecell[t]{\textbf{2,963,360}\\\textbf{($\downarrow$61.0\%)}} \\

& \makecell[t]{DSP}
& \makecell[t]{319\\[-0.2ex]~}
& \makecell[t]{319\\(0.0\%)}
& \makecell[t]{\textbf{278}\\\textbf{($\downarrow$12.9\%)}}
& \makecell[t]{319\\(0.0\%)}
& \makecell[t]{\textbf{278}\\\textbf{($\downarrow$12.9\%)}} \\

& \makecell[t]{BRAM}
& \makecell[t]{164\\[-0.2ex]~}
& \makecell[t]{91\\($\downarrow$44.5\%)}
& \makecell[t]{137\\($\downarrow$16.5\%)}
& \makecell[t]{\textbf{89}\\\textbf{($\downarrow$45.7\%)}}
& \makecell[t]{117\\($\downarrow$28.7\%)} \\

& \makecell[t]{FF}
& \makecell[t]{40,302\\[-0.2ex]~}
& \makecell[t]{42,049\\($\uparrow$4.3\%)}
& \makecell[t]{38,210\\($\downarrow$5.2\%)}
& \makecell[t]{41,997\\($\uparrow$4.2\%)}
& \makecell[t]{\textbf{38,037}\\\textbf{($\downarrow$5.6\%)}} \\

& \makecell[t]{LUT}
& \makecell[t]{81,034\\[-0.2ex]~}
& \makecell[t]{81,378\\($\uparrow$0.4\%)}
& \makecell[t]{76,360\\($\downarrow$5.8\%)}
& \makecell[t]{81,137\\($\uparrow$0.1\%)}
& \makecell[t]{\textbf{76,002}\\\textbf{($\downarrow$6.2\%)}} \\

\bottomrule
\end{tabular}%
}
\end{table}

% ============================================================
\section{Limitations and Future Work}
\label{sec:limitations}
% ============================================================

\sysname exposes three rule extraction parameters ($k$, $\alpha_{\text{sc}}$, and $\alpha_{\text{conf}}$), each with a monotone, predictable effect on a specific
hardware--accuracy trade-off (Section~\ref{sec:ablation}). A natural next step
is automated design-space exploration (DSE) that sweeps these knobs to select the Pareto-optimal
configuration for a given deployment constraint.

Bernstein activations are architecturally compatible with transformer
FFN sublayers, but rule extraction over dense latent representations yields rules with limited semantic interpretability. Concept probing~\cite{kim2018tcav} identifies semantically labeled directions in the hidden space, which could enable rule extraction over a compact, interpretable basis.

Bernstein activations are not inherently limited to tabular MLPs or transformer
FFNs. Polynomial networks have demonstrated superior representation power over
ReLU-based CNNs on vision tasks~\cite{chrysos2020p}, making convolutional
extension of \sysname's pipeline a natural future
direction.

% ============================================================
\section{Conclusion}
\label{sec:conclusion}
% ============================================================
We presented \sysname, a neurosymbolic compiler that deploys Bernstein
polynomial activations at the edge. \sysname distills large pretrained
DNNs into compact BNNs that preserve accuracy while exposing two
deployment paths: direct LUT-based FPGA synthesis and geometry-aligned
symbolic rule extraction. Across all evaluated architectures and
datasets, BNNs outperform ReLU under compression, achieving
$91.9$--$99.8$\% latency reduction and substantial resource savings over
a quantized teacher within $0.5$\,pp accuracy. The LUT-based path
eliminates the activation train--deploy gap for exact hardware
realization, while Bernstein's geometric structure yields compact,
interpretable rules with explicit input-space constraints.

\section*{Acknowledgments}
This work was supported by the National Science Foundation (NSF) under Grant No. 2504809. The authors used OpenAI ChatGPT and Anthropic Claude to assist with manuscript preparation, including editing and formatting of the text; all technical content, analysis, and conclusions are the authors' own.

\bibliographystyle{IEEEtran}
\bibliography{citations}
\vspace{-34pt}
\begin{IEEEbiography}[{\raisebox{0.6in}{\includegraphics[width=0.7in,height=0.7in,clip,keepaspectratio]{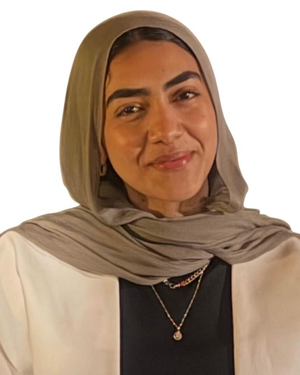}}}]{Malak Gamal El-Din}
is a Ph.D. student in the Department of Electrical Engineering and Computer Science, University of California, Irvine. Her research interests include symbolic knowledge distillation, edge AI, and multitask learning.
\end{IEEEbiography}
\vspace{-75pt}
\begin{IEEEbiography}[{\raisebox{0.6in}{\includegraphics[width=0.7in,height=0.7in,clip,keepaspectratio]{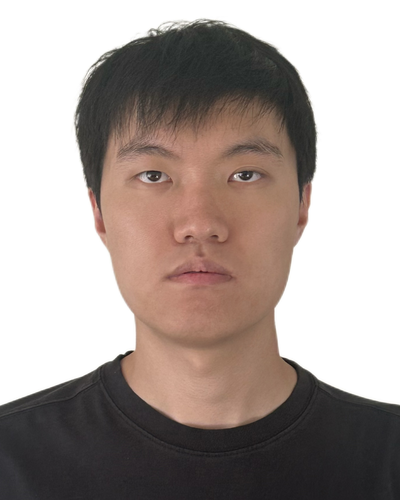}}}]
{Yifan Zhang} received the B.S. degree  from Tongji University,
Shanghai, China, in 2020. He is currently pursuing
the Ph.D. degree at University of California, Irvine, with research interests in software/hardware co-design for AI accelerators.
\end{IEEEbiography}
\vspace{-75pt}
\begin{IEEEbiography}[{\raisebox{0.6in}{\includegraphics[width=0.7in,height=0.7in,clip,keepaspectratio]{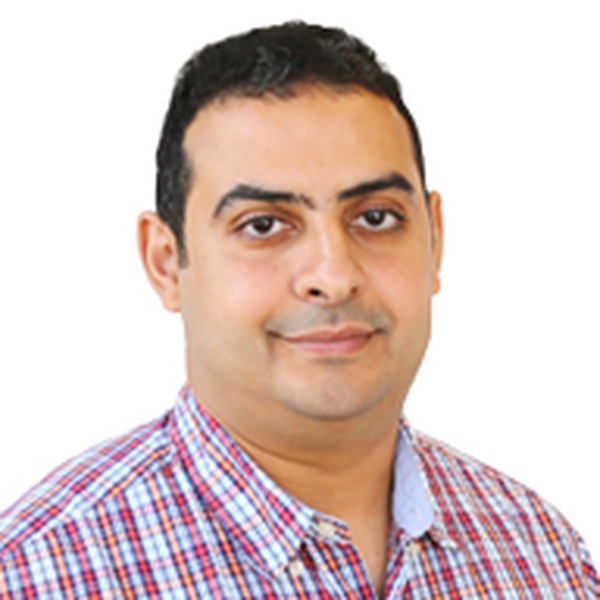}}}]{Yasser Shoukry}is an Associate Professor in the Department of Electrical Engineering and Computer Science, University of California, Irvine. His research interests include resilience and safety of AI and cyber-physical systems.
\end{IEEEbiography}
\vspace{-75pt}
\begin{IEEEbiography}[{\raisebox{0.6in}{\includegraphics[width=0.7in,height=0.7in,clip,keepaspectratio]{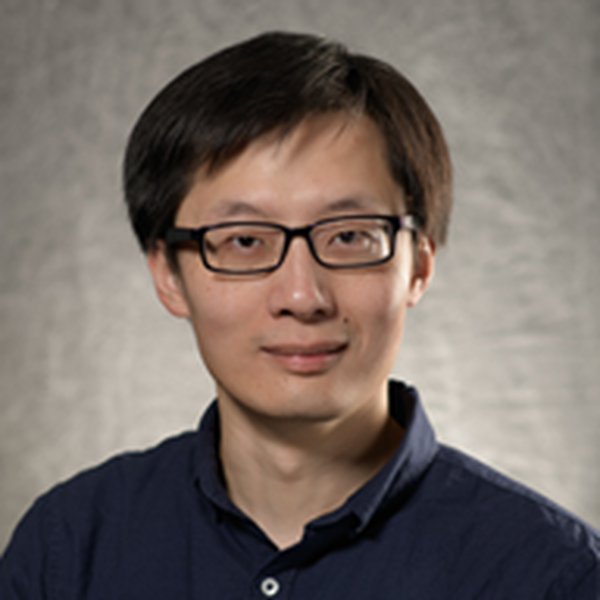}}}]{Sitao Huang}is an Assistant Professor in the Department of Electrical Engineering and Computer Science, University of California, Irvine. His research interests include hardware accelerators and programming languages for hardware systems.
\end{IEEEbiography}
\vspace{-75pt}
\begin{IEEEbiography}[{\raisebox{0.6in}{\includegraphics[width=0.7in,height=0.7in,clip,keepaspectratio]{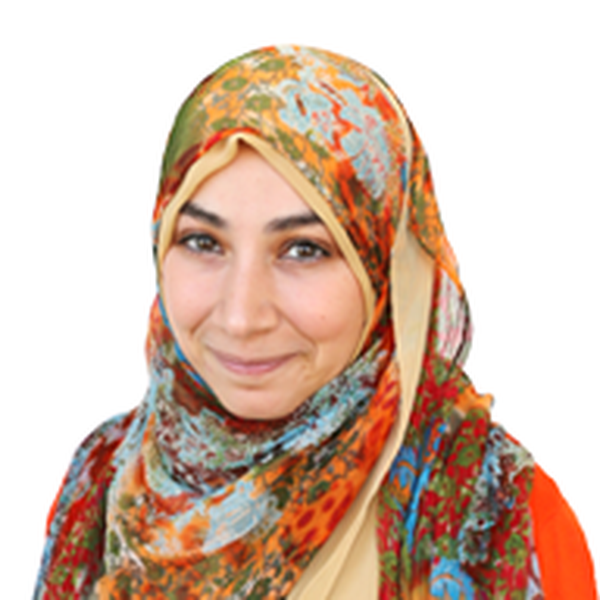}}}]{Salma Elmalaki}is an Associate Professor in the Department of Electrical Engineering and Computer Science, University of California, Irvine. Her research focuses on cyber-physical systems, human-in-the-loop systems, mobile computing, and edge AI.
\end{IEEEbiography}

\end{document}